\documentclass[sigconf]{aamas} 

\usepackage{balance} 
\usepackage[multidot]{grffile}
\usepackage{graphicx}

\usepackage{wrapfig}
\usepackage{amsthm}
\usepackage{subfig}
\usepackage{amsmath}
\usepackage[ruled]{algorithm2e}
\usepackage{float}
\usepackage{natbib}
\usepackage{mathtools}
\mathtoolsset{showonlyrefs}

\newcounter{hypo_counter}
\newcounter{scn_counter}
\newtheorem{hypothesis}[hypo_counter]{Hypothesis}
\newtheorem{scenario}[scn_counter]{Scenario}

\newcommand{\ns}{{|\mathcal{S}^+|}}
\newcommand{\E}{\mathbb{E}}
\newcommand{\R}{\mathbb{R}}

\setcopyright{ifaamas}
\acmConference[AAMAS '22]{Proc.\@ of the 21st International Conference
on Autonomous Agents and Multiagent Systems (AAMAS 2022)}{May 9--13, 2022}
{Online}{P.~Faliszewski, V.~Mascardi, C.~Pelachaud,
M.E.~Taylor (eds.)}
\copyrightyear{2022}
\acmYear{2022}
\acmDOI{}
\acmPrice{}
\acmISBN{}

\acmSubmissionID{210}

\title[Discounting Mismatch in Actor-Critic Algorithms]{A Deeper Look at Discounting Mismatch \\in Actor-Critic Algorithms}

\author{Shangtong Zhang}
\affiliation{
  \institution{University of Oxford}
  \city{}
  \country{}
  }
\email{shangtong.zhang@cs.ox.ac.uk}

\author{Romain Laroche}
\affiliation{
  \institution{Microsoft Research Montreal}
  \city{}
  \state{}
  \country{}}
\email{romain.laroche@microsoft.com}

\author{Harm van Seijen}
\affiliation{
  \institution{Microsoft Research Montreal}
  \city{}
  \state{}
  \country{}}
\email{harm.vanseijen@microsoft.com}

\author{Shimon Whiteson}
\affiliation{
  \institution{University of Oxford}
  \city{}
  \country{}
  }
\email{shimon.whiteson@cs.ox.ac.uk}

\author{Remi Tachet des Combes}
\affiliation{
  \institution{Microsoft Research Montreal}
  \city{}
  \state{}
  \country{}}
\email{remi.tachet@microsoft.com}

\begin{abstract}
We investigate the discounting mismatch in actor-critic algorithm implementations from a representation learning perspective. 
Theoretically, actor-critic algorithms usually have discounting for both the actor and critic,
\textit{i.e.}, there is a $\gamma^t$ term in the actor update for the transition observed at time $t$ in a trajectory and the critic is a discounted value function.
Practitioners, however, usually ignore the discounting ($\gamma^t$) for the actor while using a discounted critic.
We investigate this mismatch in two scenarios.
In the first scenario, 
we consider optimizing an undiscounted objective $(\gamma = 1)$ where $\gamma^t$ disappears naturally $(1^t = 1)$. 
We then propose to interpret the discounting in the critic in terms of a \emph{bias-variance-representation} trade-off and provide supporting empirical results.
In the second scenario,
we consider optimizing a discounted objective ($\gamma < 1$) and propose to interpret the omission of the discounting in the actor update
from an \emph{auxiliary task} perspective and provide supporting empirical results.
\end{abstract}

\keywords{Actor Critic; Discount Factor; Reinforcement Learning}

\begin{document}


\pagestyle{fancy}
\fancyhead{}


\maketitle 


\section{Introduction}

Actor-critic algorithms have enjoyed great success both theoretically (\citet{williams1992simple,sutton2000policy,konda2002thesis,schulman2015trust}) and empirically (\citet{mnih2016asynchronous,silver2016mastering,schulman2017proximal,opeai2018}).
There is, however, a longstanding gap between the theory behind actor-critic algorithms and how practitioners implement them.
Let $\gamma, \gamma_{\textsc{a}}$, and $\gamma_{\textsc{c}}$ be the discount factors for defining the objective,
updating the actor, and updating the critic respectively.
Theoretically, 
no matter whether $\gamma = 1$ or $\gamma < 1$,
we should always use $\gamma_{\textsc{a}} = \gamma_{\textsc{c}} = \gamma$ (\citet{sutton2000policy,schulman2015trust}) or at least keep $\gamma_{\textsc{a}} = \gamma_{\textsc{c}}$ if Blackwell optimality (\citet{veinott1969discrete,weitzman2001gamma})\footnote{Blackwell optimality states that, in finite MDPs, there exists a $\gamma_0<1$ such that for all $\gamma \geq \gamma_0$,
the optimal policies for the $\gamma$-discounted objective are the same.
} is considered.
Practitioners, however, usually use $\gamma_{\textsc{a}} = 1$ and $\gamma_{\textsc{c}} < 1$ in their implementations
(\citet{baselines,caspi_itai_2017_1134899,deeprl,pytorchrl,spinningup2018,liang2018rllib,stooke2019rlpyt}). 
Although this mismatch and its theoretical disadvantage 
have been recognized by \citet{thomas2014bias,DBLP:journals/corr/abs-1906-07073},
whether and why it yields benefits in practice 
has not been systematically studied.
In this paper,
we empirically investigate this mismatch from a representation learning perspective.
We consider two scenarios separately.

\begin{scenario}
The true objective is undiscounted ($\gamma = 1$)
\end{scenario}


The theory prescribes to use $\gamma_{\textsc{a}} = \gamma_{\textsc{c}} = \gamma = 1$. Practitioners, however, usually use $\gamma_{\textsc{a}} = \gamma = 1$ but $\gamma_{\textsc{c}} < 1$, introducing \emph{bias}. 
We explain the benefits from this mismatch with the following hypothesis:
\begin{hypothesis}
\label{hyp:undiscounted}
$\gamma_\textsc{c} < 1$ optimizes a bias-variance-representation trade-off.
\end{hypothesis}
It is easy to see that $\gamma_\textsc{c} < 1$ reduces the variance in bootstrapping targets.
We also provide empirical evidence showing that
when $\gamma_\textsc{c} < 1$, 
it may become easier
to find a good representation than when $\gamma_\textsc{c} = 1$
for reasons beyond the reduced variance.
Consequently, although using $\gamma_{\textsc{c}} < 1$ introduces bias, it can facilitate representation learning. 
For our empirical study, we make use of fixed horizon temporal difference learning (\citet{de2019fixed}) to disentangle the various effects of the discount factor on the learning process.
\begin{scenario}
    The true objective function is discounted: $\gamma=\gamma_{\textsc{c}}<1$.
\end{scenario}


Theoretically, there is a $\gamma^t$ term for the actor update on a transition observed at time $t$ in a trajectory (\citet{sutton2000policy,schulman2015trust}). Practitioners, however, usually ignore this term while using a discounted critic, \textit{i.e.}, $\gamma_{\textsc{a}} = 1$ and $\gamma_{\textsc{c}} = \gamma < 1$ are used. 
We explain this mismatch with the following hypothesis:
\begin{hypothesis}
\label{hyp:discounted}
The possible performance improvement of the biased setup (i.e., $\gamma_\textsc{c} = \gamma < 1$ and $\gamma_\textsc{a} = 1$) over the unbiased setup (i.e., $\gamma_\textsc{c} = \gamma_\textsc{a} = \gamma < 1$)
comes from improved representation learning.
\end{hypothesis}
Our empirical study involves implementing the 
difference between the biased and unbiased setup as an auxiliary task such that the difference contributes to the learning process through only representation learning.  
We also design new benchmarking environments where the sign of the reward function is flipped after a certain time step such that later transitions differ from earlier ones. 
In that setting, 
the unbiased setup outperforms the biased setup.

\section{Background}
\label{sec bg}


\textbf{Markov Decision Processes: } We consider an infinite horizon MDP with a finite state space $\mathcal{S}$, 
a finite action space $\mathcal{A}$, 
a bounded reward function $r: \mathcal{S} \to \R$,
a transition kernel $p: \mathcal{S} \times \mathcal{S} \times \mathcal{A} \to [0, 1]$,
an initial state distribution $\mu_0$,
and a discount factor $\gamma \in [0, 1]$.\footnote{Following \citet{schulman2015trust}, 
we consider $r: \mathcal{S} \to \R$ instead of $r: \mathcal{S} \times \mathcal{A} \to \R$ for simplicity.}
The initial state $S_0$ is sampled from $\mu_0$.
At time step $t$, 
an agent in state $S_t$ takes action $A_t \sim \pi(\cdot | S_t)$,
where $\pi: \mathcal{A} \times \mathcal{S} \to [0, 1]$ is the policy it follows.
The agent then gets a reward $R_{t+1} \doteq r(S_t)$ and proceeds to the next state $S_{t+1} \sim p(\cdot | S_t, A_t)$.
The return of the policy $\pi$ at time step $t$ is defined as 
\begin{align}
  \textstyle G_t \doteq \sum_{i=1}^\infty \gamma^{i-1} R_{t+i},
\end{align}
which allows us to define the state value function
\begin{align}
  v_\pi^\gamma(s) \doteq \E[G_t | S_t = s]
\end{align}
and the state-action value function
\begin{align}
  q_\pi^\gamma(s, a) \doteq \E[G_t | S_t = s, A_t = a].
\end{align}
We consider episodic tasks where we assume there is an absorbing state $s^\infty \in \mathcal{S}$ such that $r(s^\infty) = 0$ and $p(s^\infty | s^\infty, a) = 1$ holds for any $a \in \mathcal{A}$.
When $\gamma < 1$, $v_\pi^\gamma$ and $q_\pi^\gamma$ are always well defined.
When $\gamma = 1$, to ensure $v_\pi^\gamma$ and $q_\pi^\gamma$ are well defined, 
we further assume finite expected episode length.
Let $T_s^\pi$ be a random variable denoting the first time step that an agent hits $s^\infty$ when following $\pi$ given $S_0 = s$.
We assume $T_{\max} \doteq \sup_{\pi \in \Pi} \max_s \E[T_s^\pi] < \infty$,
where $\pi$ is parameterized by $\theta$ and $\Pi$ is the corresponding function class.
Similar assumptions are also used in stochastic shortest path problems (\textit{e.g.}, Section 2.2 of \citet{bertsekas1996neuro}).
In our experiments, 
all the environments have a hard time limit of $1000$,
\textit{i.e.}, $T_{\max} = 1000$. This is standard practice; classic RL environments also have an upper limit on their episode lengths (\textit{e.g.} 27k in \citet[ALE]{bellemare13arcade}).
Following \citet{pardo2018time},
we add the (normalized) time step $t$ in the state to keep the environment Markovian under the presence of the hard time limit.
We measure the performance of a policy $\pi$ with
\begin{align}
  J_\gamma(\pi) \doteq \E_{S_0 \sim \mu_0}[v_\pi^\gamma(S_0)].
\end{align}

\textbf{Vanilla Policy Gradient: } \citet{sutton2000policy} compute $\nabla_\theta J_\gamma(\pi)$ as
\begin{align}
\label{eq:pg}
\textstyle{\nabla_\theta J_\gamma(\pi) \doteq \sum_{s } d_\pi^\gamma(s) \sum_{a } q_\pi^\gamma(s, a) \nabla_\theta \pi(a|s)},
\end{align}
where 
\begin{align}
  d_\pi^\gamma(s) \doteq \begin{cases}
\sum_{t=0}^\infty \gamma^t \Pr(S_t = s | \mu_0, p, \pi), & \gamma < 1 \\
\E[\sum_{t=0}^{T_{S_0}^\pi} \Pr(S_t = s | S_0, p, \pi)], &\gamma = 1
  \end{cases}.
\end{align}
Note $d_\pi^\gamma$ remains well-defined for $\gamma = 1$ when $T_{\max} < \infty$. 
In order to optimize the policy performance $J_\gamma(\pi)$, one can follow \eqref{eq:pg} and,
at time step $t$, 
update $\theta_t$ as
\begin{align}
\label{eq:pg_update}
\theta_{t+1} \gets \theta_t + \alpha \gamma_{\textsc{a}}^t q_\pi^{\gamma_{\textsc{c}}}(S_t, A_t) \nabla_\theta \log \pi(A_t | S_t),
\end{align}
where $\alpha$ is a learning rate.
If we replace $q_\pi^{\gamma_{\textsc{c}}}$ with a learned value function,
the update rule \eqref{eq:pg_update} becomes an actor-critic algorithm,
where the actor refers to $\pi$ and the critic refers to the learned approximation of $q_\pi^{\gamma_{\textsc{c}}}$.
In practice,
an estimate for $v_\pi^{\gamma_{\textsc{c}}}$
instead of $q_\pi^{\gamma_{\textsc{c}}}$ is usually learned.
Theoretically,
we should have $\gamma_{\textsc{a}} = \gamma_{\textsc{c}} = \gamma$.
Practitioners, however, usually ignore the $\gamma_{\textsc{a}}^t$ term in \eqref{eq:pg_update},
and
use $\gamma_{\textsc{c}} < \gamma_{\textsc{a}} = 1$. What this update truly optimizes remains an open problem~(\citet{DBLP:journals/corr/abs-1906-07073}).

\textbf{TRPO and PPO: } To improve the stability of actor-critic algorithms,
\citet{schulman2015trust} propose Trust Region Policy Optimization (TRPO), 
based on the performance improvement lemma:
\begin{lemma}
\label{lem:trpo}
(Theorem 1 in \citet{schulman2015trust}) For $\gamma < 1$ and any two policies $\pi$ and $\pi^\prime$,
\begin{align*}
J_\gamma(\pi^\prime) \geq& J_\gamma(\pi) + \textstyle{\sum_s d_\pi^\gamma(s) \sum_a \pi^\prime(a | s) {\text{Adv}}_\pi^\gamma(s, a)} \\
&\textstyle{- \frac{4 \max_{s,a} |{\text{Adv}}_\pi^\gamma(s, a)| \gamma \epsilon(\pi, \pi^\prime) }{(1 - \gamma)^2}},
\end{align*}
where 
\begin{align}
  {\text{Adv}}_\pi^\gamma(s, a) \doteq \E_{s^\prime \sim p(\cdot | s, a)}[r(s) + \gamma v_\pi^\gamma(s^\prime) - v_\pi^\gamma(s)]
\end{align}
is the advantage,
\begin{align}
  \epsilon(\pi, \pi^\prime) \doteq \max_s D_{\text{KL}}(\pi(\cdot | s) || \pi^\prime(\cdot | s)),
\end{align}
and $D_{\text{KL}}$ refers to the KL divergence.
\end{lemma}
To facilitate our empirical study, 
we first make a theoretical contribution by extending Lemma~\ref{lem:trpo} to the undiscounted setting. 
We have the following lemma:
\begin{lemma}
\label{lem:trpo-undiscounted}
Assuming $T_{\max} < \infty$, for $\gamma = 1$ and any two policies $\pi$ and $\pi^\prime$,
we have
\begin{align*}
J_\gamma(\pi^\prime) \geq& \textstyle{J_\gamma(\pi) + \sum_s d_\pi^\gamma(s) \sum_a \pi^\prime(a | s) {\text{Adv}}_\pi^\gamma(s, a)} \\
&\textstyle{- 4 \max_{s,a} |{\text{Adv}}_\pi^\gamma(s, a)| T^2_{\max} \epsilon(\pi, \pi^\prime)}.
\end{align*}
\end{lemma}
The proof of Lemma~\ref{lem:trpo-undiscounted} is provided in the appendix. 
A practical implementation of Lemmas~\ref{lem:trpo} and~\ref{lem:trpo-undiscounted}
is to compute a new policy $\theta$ via gradient ascent on the clipped objective:
\begin{align}
\label{eq:ppo-loss}
L(\theta) \doteq& \textstyle \sum_{t=0}^{\infty} \gamma_\textsc{a}^t L_t(\theta, \theta_\text{old}), 
\end{align}
where
\begin{align}
\textstyle L_t(\theta, \theta_\text{old}) \doteq& \textstyle \min \Big\{\frac{\pi_{\theta}(A_t | S_t)}{\pi_{\theta_\text{old}}(A_t | S_t)} \text{Adv}^{\gamma_{\textsc{c}}}_{\pi_{\theta_\text{old}}}(S_t, A_t), \\
&\text{clip}(\frac{\pi_{\theta}(A_t | S_t)}{\pi_{\theta_\text{old}}(A_t|S_t)}) \text{Adv}^{\gamma_{\textsc{c}}}_{\pi_{\theta_\text{old}}}(S_t, A_t) \Big\} \nonumber.
\end{align}
Here $S_t$ and $A_t$ are sampled from $\pi_{\theta_\text{old}}$,
and
\begin{align}
  \text{clip}(x) \doteq \max(\min(x, 1 + \epsilon), 1 - \epsilon)
\end{align}
with $\epsilon$ a hyperparameter.
Theoretically,
we should have $\gamma_{\textsc{a}} = \gamma_{\textsc{c}}$, but
practical algorithms like Proximal Policy Optimization (PPO, \citet{schulman2017proximal})
usually use $\gamma_{\textsc{c}} < \gamma_{\textsc{a}} = 1$.

\textbf{Policy Evaluation: } We now introduce several policy evaluation techniques we use in our empirical study.
Let $\hat{v}$ be our estimate of $v_\pi^\gamma$.
At time step $t$,
Temporal Difference learning (TD, \citet{sutton1988learning}) updates $\hat{v}$ as 
\begin{align}
  \hat{v}(S_t) \gets \hat{v}(S_t) + \alpha (R_{t+1} + \gamma \hat{v}(S_{t+1}) - \hat{v}(S_t)).
\end{align}
Instead of the infinite horizon discounted return $G_t$,
\citet{de2019fixed} propose to consider the $H$-step return
\begin{align}
  \textstyle G_t^H \doteq \sum_{i=1}^H R_{t+i}.
\end{align}
Correspondingly,
the $H$-step value function is defined as
\begin{align}
  v_\pi^H(s) \doteq \E[G_t^H | S_t = s].
\end{align}
We let $\hat{v}^H$ be our estimate of $v_\pi^H$.
At time step $t$,
\citet{de2019fixed} use the following update rule to learn $\hat{v}^H$. For $i = 1, \dots, H$:
\begin{align}
\label{eq:fhtd}
\hat{v}^i(S_t) \gets \hat{v}^i(S_t) + \alpha (R_{t+1} + \hat{v}^{i-1}(S_{t+1}) - \hat{v}^i(S_t)),
\end{align}
where $\hat{v}^0(s) \doteq 0$.
In other words,
to learn $\hat{v}^H$, 
we need to learn $\{\hat{v}^i\}_{i = 1, \dots, H}$ simultaneously.
\citet{de2019fixed} call \eqref{eq:fhtd}  \textit{Fixed Horizon Temporal Difference} learning (FHTD).

\begin{figure*}[t]
\centering
\includegraphics[width=\linewidth]{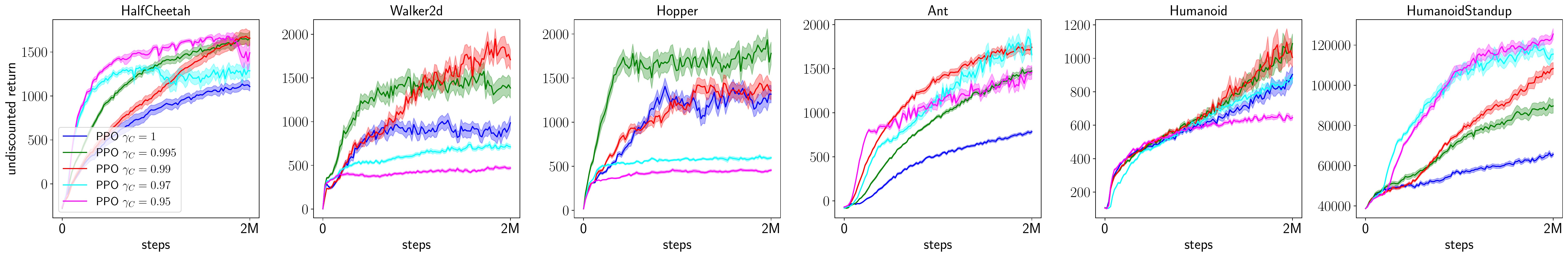}
\caption{\label{fig:ppo-ret} The default PPO implementation with different discount factors.}
\end{figure*}

\begin{figure*}[t]
\centering
\includegraphics[width=\linewidth]{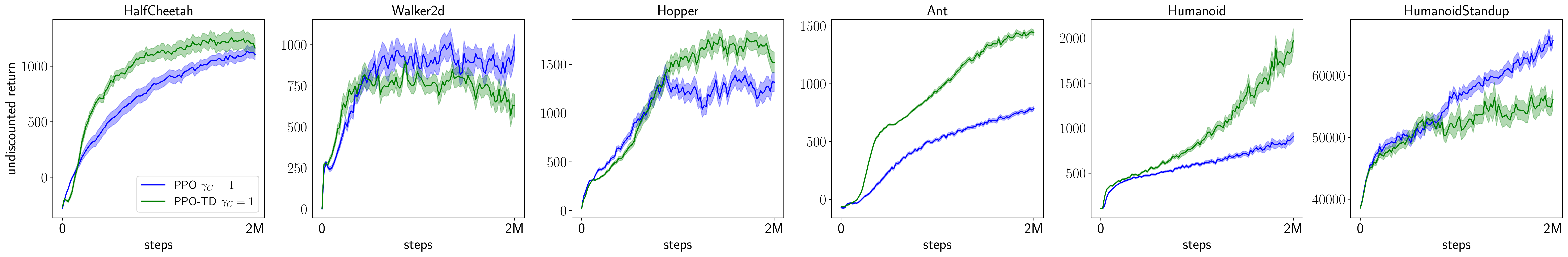}
\caption{\label{fig:ppo-td-mc-ret} Comparison between PPO and PPO-TD when $\gamma_{\textsc{c}} = 1$.}
\end{figure*}

\textbf{Methodology: }
We consider MuJoCo (\citet{todorov2012mujoco}) robot simulation tasks from OpenAI gym (\citet{brockman2016openai}) as our benchmark.
Given its popularity in understanding deep RL algorithms~(\citet{henderson2017deep,ilyas2018closer,engstrom2019implementation,andrychowicz2020matters})
and designing new deep RL algorithms (\citet{fujimoto2018addressing,haarnoja2018soft}),
we believe our empirical results are relevant to most practitioners.

We choose PPO, a simple yet effective and widely used algorithm, as the representative actor-critic algorithm for our empirical study.
PPO is usually equipped with generalized advantage estimation~(GAE, \citet{schulman2015high}),
which has a tunable hyperparameter $\hat{\gamma}$.
The roles of $\gamma$ and $\hat{\gamma}$ are similar.
To reduce its confounding effect, 
we do not use GAE in our experiments,
\textit{i.e.},
the advantage estimation for our actor is simply the TD error $R_{t+1} + \gamma_{\textsc{c}} \hat{v}(S_{t+1}) - \hat{v}(S_t)$.
The PPO pseudocode we follow is provided in Algorithm~\ref{algo:ppo} in the appendix
and we refer to it as the default PPO implementation.

We use the standard architecture and optimizer across all tasks.
In particular, the actor and the critic do not share layers.
We conduct a thorough grid search for the learning rate of each algorithmic configuration (\textit{i.e.}, for every curve in all figures).
All experimental details are provided in the appendix.
We report the average episode return of the ten most recent episodes against the number of interactions with the environment.
Curves are averages over 30 independent runs with shaded regions indicating standard errors.
All our implementations and our Docker environment are publicly available for future research.\footnote{\url{https://github.com/ShangtongZhang/DeepRL}}

\begin{figure*}[t]
\centering
\includegraphics[width=\linewidth]{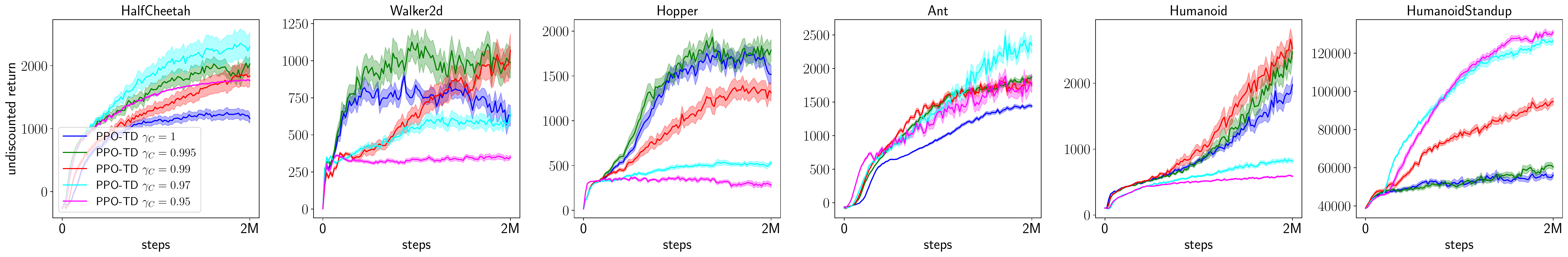}
\caption{\label{fig:ppo-td-critic-ret} PPO-TD with different discount factors.}
\end{figure*}

\begin{figure*}[t]
\centering
\includegraphics[width=\linewidth]{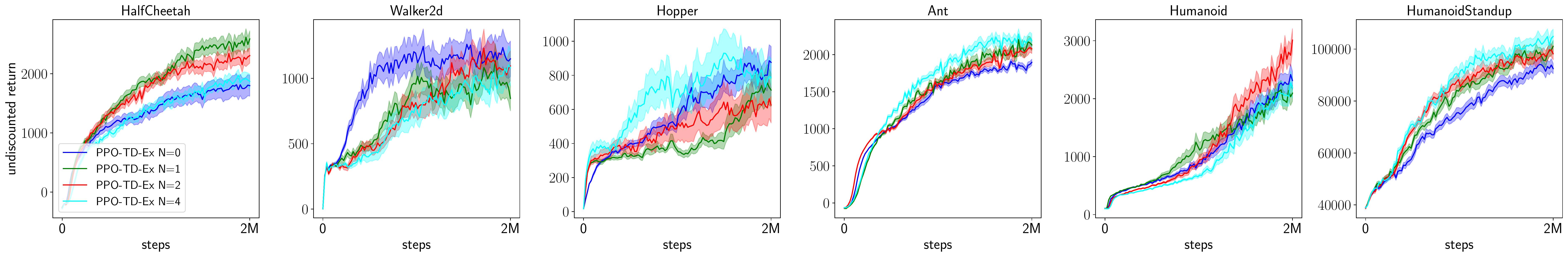}
\caption{\label{fig:ppo-td-mp-ret-0.99} PPO-TD-Ex ($\gamma_{\textsc{c}} = 0.99$).}
\end{figure*}
\begin{figure*}[t]
\centering
\includegraphics[width=\linewidth]{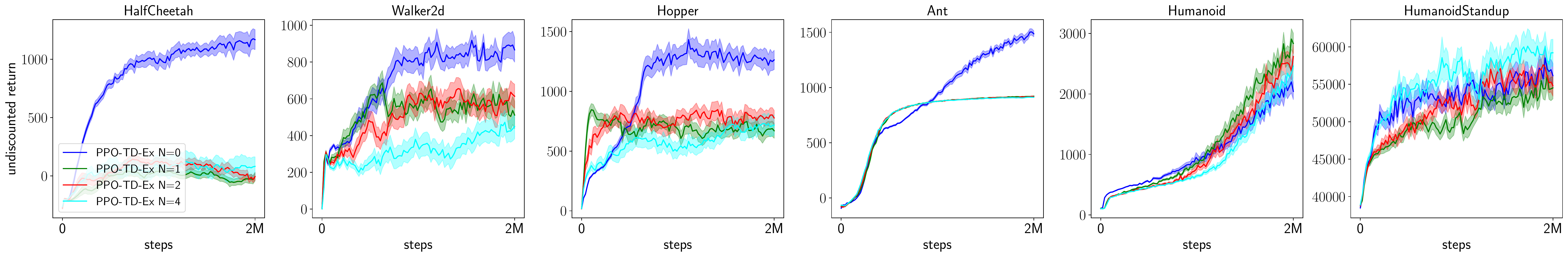}
\caption{\label{fig:ppo-td-mp-ret-1} PPO-TD-Ex ($\gamma_{\textsc{c}} = 1$).}
\end{figure*}

\section{Optimizing the Undiscounted Objective (Scenario~\ref{hyp:undiscounted})}
\label{sec undiscounted}
In this scenario, 
the goal is to optimize the \emph{undiscounted} objective $J_{\gamma = 1}(\pi)$.
This scenario is related to most practitioners as they usually use the undiscounted return as the performance metric (\citet{mnih2016asynchronous,schulman2017proximal,haarnoja2018soft}). 
One theoretically grounded option is to use $\gamma_{\textsc{a}} = \gamma_{\textsc{c}} = \gamma = 1$.
By using $\gamma_{\textsc{a}} = 1$ and $\gamma_{\textsc{c}} < 1$,
practitioners introduce \emph{bias}.
We first empirically confirm that introducing bias in this way indeed has empirical advantages.
A simple first hypothesis for this is that $\gamma_{\textsc{c}} < 1$ leads to lower variance in Monte Carlo return bootstrapping targets than $\gamma_{\textsc{c}} = 1$; it thus optimizes a bias-variance trade-off.
However,
we further show that there are empirical advantages from $\gamma_{\textsc{c}} < 1$ that cannot be explained solely by this bias-variance trade-off,
indicating that there are additional factors beyond variance.
We then show empirical evidence identifying representation learning as an additional factor,
leading to the \emph{bias-variance-representation} trade-off from Hypothesis~\ref{hyp:undiscounted}.
All the experiments in this section use $\gamma_{\textsc{a}} = 1$.

\textbf{Bias-variance trade-off:} To investigate the advantages of using $\gamma_{\textsc{c}} < 1$,
we first test default PPO with $\gamma_{\textsc{c}} \in \{0.95, 0.97, 0.99, 0.995, 1\}$.
We find that the best discount factor is always with $\gamma_{\textsc{c}} < 1$ and that $\gamma_{\textsc{c}} = 1$ usually leads to a performance drop (Figure~\ref{fig:ppo-ret}).
In default PPO,
although the advantage is computed as the one-step TD error,
the update target for updating the critic $\hat{v}(S_t)$ is almost always a Monte Carlo return,
\emph{i.e.}, $\sum_{i={t+1}}^{T_{\max}} \gamma^{i - t - 1}_\textsc{c} R_i$.
Here $\gamma_\textsc{c}$ has a gating effect in controlling the variance of the Monte Carlo return:
when $\gamma_\textsc{c}$ is smaller, 
the randomness from $R_i$ contributes less to the variance of the Monte Carlo return.
In this paper,
we refer to this gating effect as \emph{variance control}.
As the objective is undiscounted and we use $\gamma_{\textsc{a}} = \gamma = 1$,
theoretically we should also use $\gamma_{\textsc{c}} = 1$ when computing the Monte Carlo return if we do not want to introduce bias.
By using $\gamma_\textsc{c} < 1$,
bias is introduced.
The variance of the Monte Carlo return is, however, also reduced.
Consequently, 
a simple hypothesis for the empirical advantage of using $\gamma_{\textsc{c}} < 1$ is that it optimizes a bias-variance trade-off. 
We find, however, that there is more at play.

\textbf{Beyond bias-variance trade-off:} 
To make other possible effects of $\gamma_\textsc{c}$ pronounced,
it is desirable to reduce the variance control effect of $\gamma_\textsc{c}$.
To this end,
we benchmark PPO-TD (Algorithm~\ref{algo:ppo-td} in the appendix). 
PPO-TD is the same as default PPO except that the critic is updated with one-step TD,
\textit{i.e.},
the update target for $\hat{v}(S_t)$ is now $R_{t+1} + \gamma_{\textsc{c}} \hat{v}(S_{t+1})$.
In this update target, $\gamma_\textsc{c}$ gates the randomness from only the immediate successor state $S_{t+1}$.
By contrast,
in the original Monte Carlo update target,
$\gamma_\textsc{c}$ gates the randomness of all future states and rewards.
Figure~\ref{fig:ppo-td-mc-ret} shows that PPO-TD $(\gamma_{\textsc{c}} = 1)$ outperforms PPO $(\gamma_{\textsc{c}} = 1)$ in four games.
This indicates that PPO-TD might be less vulnerable to the large variance in critic update targets introduced by using $\gamma_\textsc{c} = 1$ than default PPO.
Figure~\ref{fig:ppo-td-critic-ret} suggests, however,
that even for PPO-TD,
$\gamma_{\textsc{c}} < 1$ is still preferable to $\gamma_{\textsc{c}} = 1$.

\begin{figure*}[t]
\centering
\includegraphics[width=\linewidth]{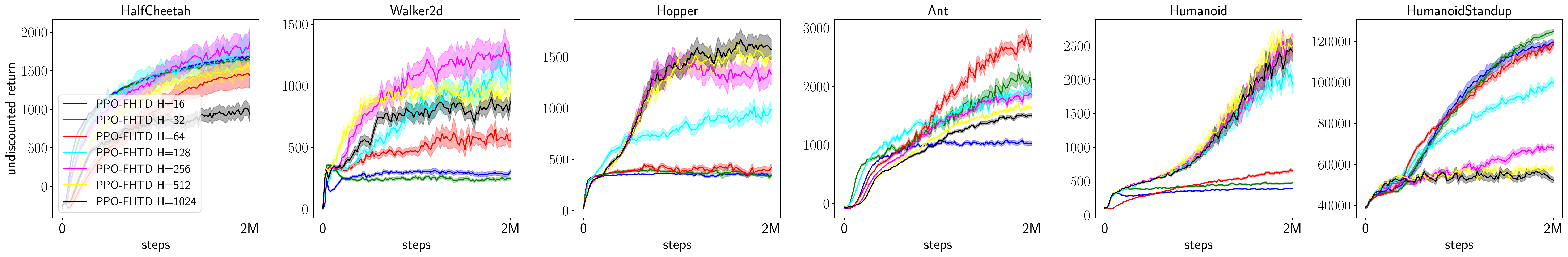}
\caption{\label{fig:ppo-fhtd-vs-ppo-td-ret} PPO-FHTD with the first parameterization. The best $H$ and $\gamma_{\textsc{c}}$ are used for each game.}
\end{figure*}
\begin{figure*}[t]
\centering
\includegraphics[width=\linewidth]{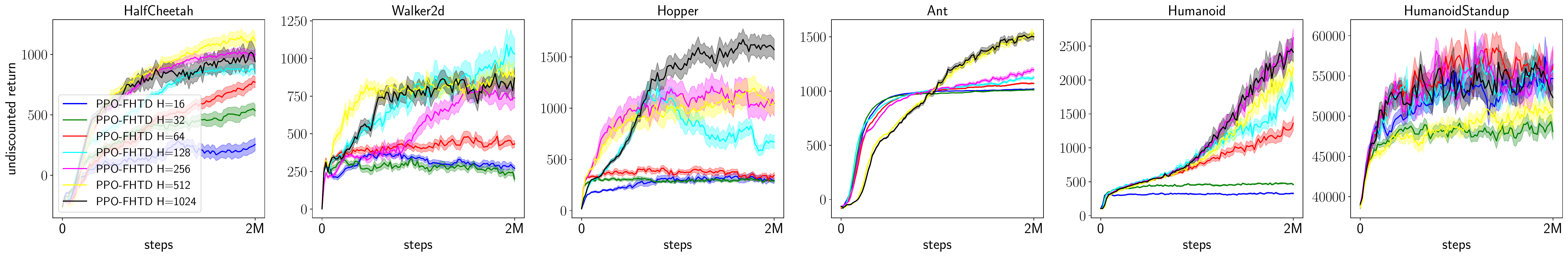}
\caption{\label{fig:ppo-active-fhtd-ret} PPO-FHTD with the second parameterization.}
\end{figure*}

Of course,
in PPO-TD, $\gamma_\textsc{c}$ still has the variance control effect,
though not as pronounced as that in default PPO.
To make other possible effects of $\gamma_\textsc{c}$ more pronounced,
we benchmark PPO-TD-Ex (Algorithm~\ref{algo:ppo-td-ex} in the appendix),
in which we provide $N$ extra transitions to the critic by sampling multiple actions at any single state and using an averaged bootstrapping target.
The update target for $\hat{v}(S_t)$ in PPO-TD-Ex is 
\begin{align}
  \textstyle \frac{1}{N + 1}\sum_{i=0}^N R_{t+1}^i + \gamma_{\textsc{c}} \hat{v}(S_{t+1}^i).
\end{align}
Here $R_{t+1}^0$ and $S_{t+1}^0$ refer to the original reward and successor state.
To get $R_{t+1}^i$ and $S_{t+1}^i$ for $i \in \{1, \dots, N\}$,
we first sample an action $A_t^i$ from the sampling policy, then reset the environment to $S_t$, and finally execute $A_t^i$ to get $R_{t+1}^i$ and $S_{t+1}^i$.
The advantage for the actor update in PPO-TD-Ex is estimated with $R_{t+1}^0 + \hat{v}(S_{t+1}^0) - \hat{v}(S_t)$ regardless of $\gamma_{\textsc{c}}$ to further depress its variance control effect.
Importantly, we do not count those $N$ extra transitions in the $x$-axis when plotting.
If we use the true value function instead of $\hat{v}$, 
$N \geq 1$ should always outperform $N = 0$
as the additional transitions help reduce variance (assuming $\gamma_\textsc{c}$ is fixed).
However,
in practice we have only $\hat{v}$,
which is not trained on the extra successor states $\{S_{t+1}^i\}_{i=1, \dots, N}$.
So the quality of the prediction $\hat{v}(S_{t+1}^i)$ depends mainly on the generalization of $\hat{v}$.
Consequently,
increasing $N$ risks potential erroneous prediction $\hat{v}(S_{t+1}^i)$.
That being said,
though not guaranteed to improve the performance,
when the prediction $\hat{v}(S_{t+1}^i)$ is decent,
increasing $N$ should at least not lead to a performance drop.
As shown by Figure~\ref{fig:ppo-td-mp-ret-0.99}, PPO-TD-Ex ($\gamma_{\textsc{c}} = 0.99$) roughly follows this intuition.
However,
surprisingly,
providing any extra transition this way to PPO-TD-Ex ($\gamma_{\textsc{c}} = 1$) leads to a significant performance drop in 4 out of 6 tasks (Figure~\ref{fig:ppo-td-mp-ret-1}).
This drop suggests that the quality of the critic $\hat{v}$,
at least in terms of making prediction on untrained states $\{S_{t+1}^i\}_{1, \dots, N}$,
is lower when $\gamma_{\textsc{c}} = 1$ is used than $\gamma_{\textsc{c}} < 1$.
In other words,
the generalization of $\hat{v}$ becomes poorer when $\gamma_{\textsc{c}}$ is increased from $0.99$ to $1$.
The curves for PPO-TD-Ex ($\gamma_{\textsc{c}} = 0.995$) are a mixture of $\gamma_{\textsc{c}} = 0.99$ and $\gamma_{\textsc{c}} = 1$ and are provided in Figure~\ref{fig:ppo-td-mp-ret-0.995} in the appendix.
The limited generalization could imply that representation learning becomes harder when $\gamma_\textsc{c}$ is increased.
By representation learning,
we refer to learning the lower layers (backbone) of a neural network.
The last layer of the neural network is then interpreted as a linear function approximator whose features are the output of the backbone.
This interpretation of representation learning is widely used in the RL community, see, \textit{e.g.},~\citet{jaderberg2016reinforcement,chung2018two,veeriah2019discovery}.


In PPO-TD, 
the bootstrapping target for training $\hat{v}(S_t)$ is $R_{t+1} + \gamma_\textsc{c} \hat{v}(S_{t+1})$,
where $\gamma_\textsc{c}$ has two roles.
First, it gates the randomness from $S_{t+1}$,
which is the aforementioned variance control.
Second, it affects the value function $v_\pi^{\gamma_\textsc{c}}$ that we want to approximate via changing the horizon of the policy evaluation problem,
which could possibly affect the difficulty of learning a good estimate $\hat{v}$ for $v_\pi^{\gamma_\textsc{c}}$ directly,
not through the variance,
which we refer to as \emph{learnability control} (see, e.g., \citet{lehnert2018value,laroche2018reinforcement,romoff2019separating}).
Both roles can be responsible for the increased difficulty in representation learning when $\gamma_\textsc{c}$ is increased.
In the rest of this section,
we provide empirical evidence showing that the changed difficulty in representation learning,
resulting directly from the changed horizon of the policy evaluation problem, 
is at play when using the $\gamma_\textsc{c} < 1$,
which,
together with the previously established bias-variance trade-off,
suggests that a \emph{bias-variance-representation trade-off} is at play when practitioners use $\gamma_\textsc{c} < 1$. 

\textbf{Bias-representation trade-off:} 
To further disentangle the variance control effect and learnability control effect of $\gamma_\textsc{c}$,
we use FHTD to train the critic in PPO,
which we refer to as PPO-FHTD (Algorithm~\ref{algo:ppo-fhtd} in the appendix).
PPO-FHTD always uses $\gamma_{\textsc{c}} = 1$ regardless of $H$.
The critic update target in PPO-TD is $R_{t+1} + \gamma_\textsc{c} \hat{v}(S_{t+1})$,
whose variance is 
\begin{align}
\label{eq var-ppo-td}
&Var(R_{t+1} + \gamma_\textsc{c} \hat v(S_{t+1}) | S_t) \\
=& Var(r(S_t) + \gamma_\textsc{c} \hat v(S_{t+1}) | S_t) \\
=& \gamma_\textsc{c}^2 Var(\hat v(S_{t+1}) | S_t).
\end{align}
By contrast, 
the critic update target in PPO-FHTD for $\hat v^i(S_t)$ is $R_{t+1} + \hat{v}^{i-1}(S_{t+1})$,
with variance:
\begin{align}
\label{eq var-ppo-fhtd}
Var(R_{t+1} + \hat v^{i-1}(S_{t+1}) | S_t) = Var(\hat v^{i-1}(S_{t+1}) | S_t).
\end{align}
On the one hand, 
manipulating $H$ in FHTD changes the horizon of the policy evaluation problem,
which corresponds to the role of learnability control of $\gamma_\textsc{c}$.
On the other hand,
manipulating $H$ does not change the multiplier proceeding the variance term (c.f. \eqref{eq var-ppo-td} and \eqref{eq var-ppo-fhtd}) and thus separates variance control from the learnability control.

\begin{figure}
  \centering
\subfloat[The first parameterization]{\includegraphics[width=0.5\linewidth]{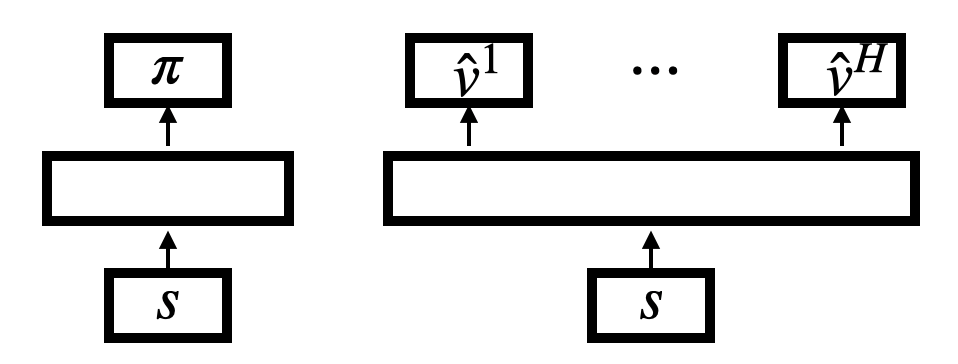}} \\
\subfloat[The second parameterization]{\includegraphics[width=0.5\linewidth]{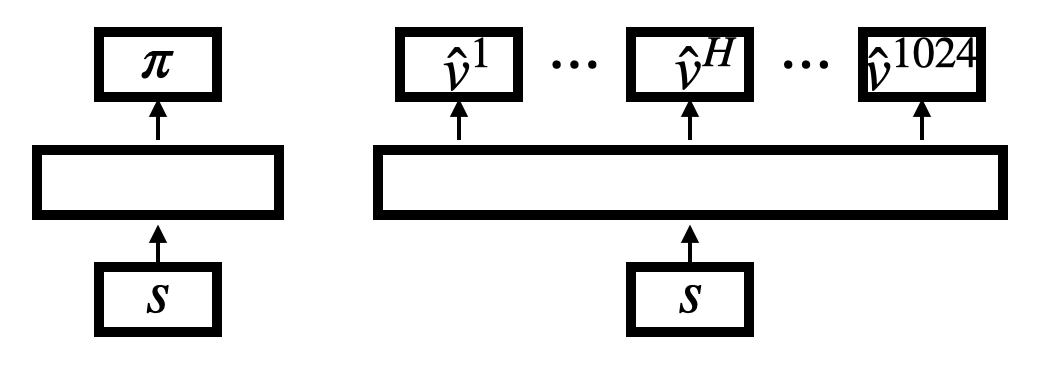}}
\caption{\label{fig:fhtd parameterization}Two parameterization of PPO-FHTD}
\end{figure}

We test two parameterizations for PPO-FHTD to investigate representation learning. 
In the first parameterization, 
to learn $v_\pi^H$, 
we parameterize $\{v_\pi^i\}_{i = 1, \dots, H}$ as $H$ different heads over the same representation layer (backbone). 
In the second parameterization,
we always learn $\{v_\pi^i\}_{i = 1, \dots, 1024}$ as 1024 different heads over the same representation layer,
regardless of what $H$ we are interested in.
To approximate $v_\pi^H$, we then simply use the output of the $H$-th head.
Figure~\ref{fig:fhtd parameterization}
further illustrates the difference between the two parameterizations. 

\begin{figure*}[t]
\centering
\includegraphics[width=0.9\linewidth]{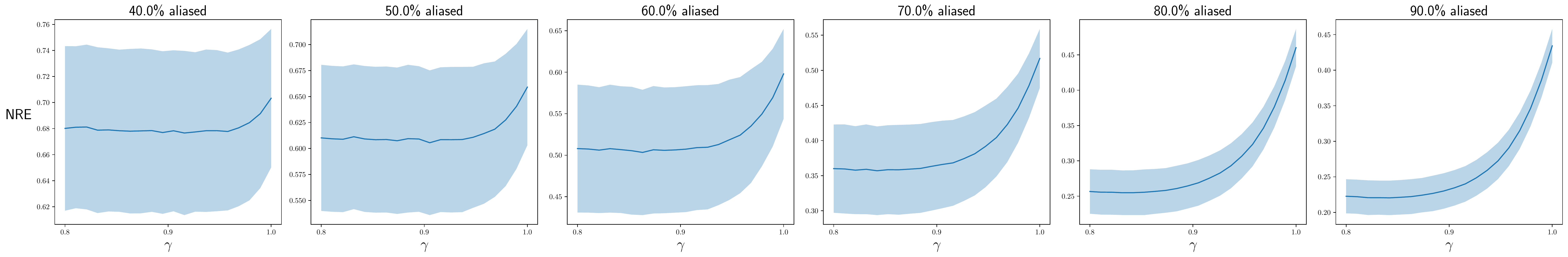}
\caption{\label{fig:state_aliasing} Normalized representation error as a function of the discount factor.
Shaded regions indicate one standard derivation.
\protect\footnotemark}
\end{figure*}
\footnotetext{The trend that NRE decreases as $\alpha$ increases is merely an artifact from how we generate $v_\gamma$.}

\begin{figure}
\centering
\includegraphics[width=0.6\linewidth]{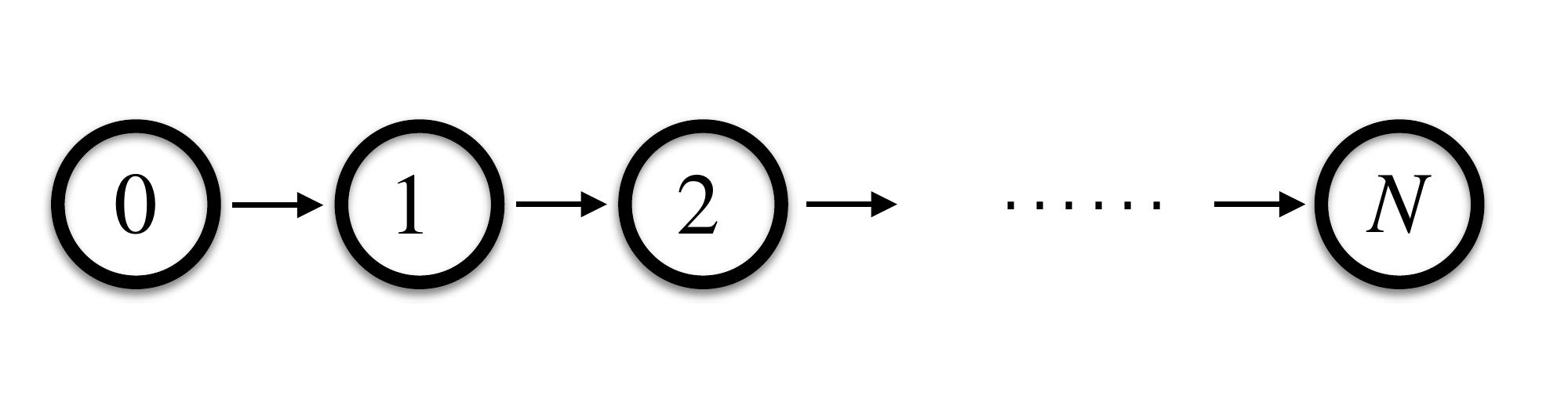}
\caption{\label{fig:chain} A simple MRP.}
\end{figure}

Figure~\ref{fig:ppo-fhtd-vs-ppo-td-ret} shows that
with the first parameterization,
the best $H$ for PPO-FHTD is usually smaller than $1024$.
Figure~\ref{fig:ppo-active-fhtd-ret},
however,
suggests that for the second parameterization, $H=1024$ is almost always among the best choices of $H$.
Comparing Figures~\ref{fig:ppo-td-critic-ret} and~\ref{fig:ppo-active-fhtd-ret}
shows that the performance of PPO-FHTD ($H=1024$) is close to the performance of PPO-TD ($\gamma_\textsc{c} = 1$)
as expected, 
since for any $H \geq T_{\max} = 1000$,
we always have $v_\pi^H(s) \equiv v_\pi^{\gamma=1}(s)$.
This performance similarity suggests that 
learning $\{v_\pi^i\}_{i = 1, \dots, 1023}$ is not an additional overhead for the network in terms of learning $v_\pi^{H=1024}$,
\textit{i.e.},
increasing $H$ does not pose additional challenges in terms of network capacity.
Then, comparing Figures~\ref{fig:ppo-fhtd-vs-ppo-td-ret} and~\ref{fig:ppo-active-fhtd-ret},
we conclude that in the tested domains, 
learning $v_\pi^H$ with different $H$  requires different representations.
This suggests that we can interpret the results in Figure~\ref{fig:ppo-fhtd-vs-ppo-td-ret} as a \emph{bias-representation trade-off}.
Using a larger $H$ is less biased but
representation learning \emph{may} become harder due to the longer policy evaluation horizon. 
Consequently,
an intermediate $H$ achieves the best performance in Figure~\ref{fig:ppo-fhtd-vs-ppo-td-ret}.
As reducing $H$ cannot bring in advantages in representation learning under the second parameterization,
the less biased $H$, \textit{i.e.}, the larger $H$ 
usually performs better in Figure~\ref{fig:ppo-active-fhtd-ret}.
Overall, $\gamma_{\textsc{c}}$ optimizes a \emph{bias-representation} trade-off by changing the policy evaluation horizon $H$.  


We further conjecture that representation learning may be harder for a longer horizon because the volume of all of good representations can become smaller.
We provide a simulated example to support this.
Consider policy evaluation on the simple Markov Reward Process (MRP) from Figure~\ref{fig:chain}.
We assume the reward for each transition is fixed,
which is randomly generated in $[0, 1]$.
Let $x_s \in \R^K$ be the feature vector for a state $s$;
we set its $i$-th component as $x_s[i] \doteq \tanh(\xi)$,
where $\xi$ is a random variable uniformly distributed in $[-2, -2]$. 
We choose this feature setup as we use $\tanh$ as the activation function in our PPO.
We use $X \in \R^{N \times K}$ to denote the feature matrix.
To create state aliasing (\citet{mccallum1997reinforcement}),
which is common under function approximation,
we first randomly split the $N$ states into $\mathcal{S}_1$ and $\mathcal{S}_2$ such that $|\mathcal{S}_1| = \alpha N$ and $|\mathcal{S}_2| = (1 - \alpha) N$,
where $\alpha$ is the proportion of states to be aliased.
Then for every $s \in \mathcal{S}_1$,
we randomly select an $\hat{s} \in \mathcal{S}_2$ and set $x_s \gets x_{\hat{s}}$.
Finally, we add Gaussian noise $\mathcal{N}(0, 0.1^2)$ to each element of $X$. 
We use $N = 100$ and $K = 30$ in our simulation and report the normalized representation error (NRE) as a function of $\gamma$.
For a feature matrix $X$,
the NRE is computed \emph{analytically} as 
\begin{align}
  \text{NRE}(\gamma) \doteq \frac{\min_w ||Xw - v_\gamma||_2}{||v_\gamma||_2},
\end{align}
where $v_\gamma$ is the \emph{analytically} computed true value function of the MRP. 
We report the results in Figure~\ref{fig:state_aliasing},
where each data point is averaged over $10^4$ randomly generated feature matrices ($X$) and reward functions.
In this MRP,
the average representation error becomes larger as $\gamma$ increases,
which suggests that,
in this MRP,
the volume of good representations (e.g., representations whose normalized representation error are smaller than some threshold) becomes smaller under a larger $\gamma$ than that under a smaller $\gamma$.

Importantly,
in this MRP experiment,
\emph{we compute all the quantities analytically so no variance in involved within a single trial}.
Consequently,
representation error is a property of $v_\gamma$ itself.
We report the unnormalized representation error in Figure~\ref{fig:state_aliasing_false} in the appendix,
where the trend is much clearer.

Overall, though we do not claim that there is a monotonic relationship between the discount factor and the difficulty of representation learning,
our empirical study suggests that representation learning is a key factor at play in the misuse of the discounting in actor-critic algorithms,
beyond the widely recognized bias-variance trade-off. 

\section{Optimizing the Discounted Objective (Scenario~\ref{hyp:discounted})}
\label{sec discounted}
\begin{figure*}[t]
\centering
\includegraphics[width=\linewidth]{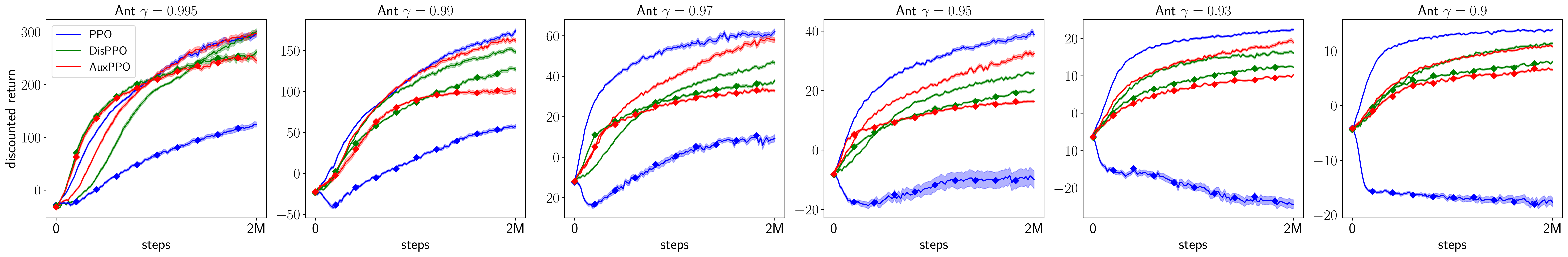}
\includegraphics[width=\linewidth]{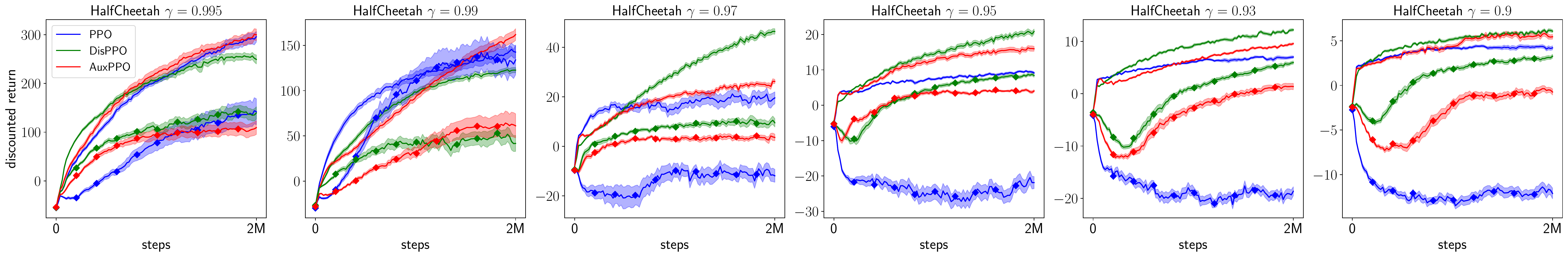}
\caption{\label{fig:aux_ppo} Curves without any marker are obtained in the original \texttt{Ant} / \texttt{HalfCheetah}.
Diamond-marked curves are obtained in \texttt{Ant} / \texttt{HalfCheetah} with $r^\prime$.
\protect\footnotemark
}
\end{figure*}
\footnotetext{See Section~\ref{sec:app_method} for more details about task selection.}
When our goal is to optimize the \emph{discounted} objective $J_{\gamma < 1}(\pi)$,
theoretically we should have the $\gamma_{\textsc{a}}^t$ term in the actor update and use $\gamma_{\textsc{c}} < 1$.
Practitioners, however, usually ignore this $\gamma_{\textsc{a}}^t$ (\textit{i.e.}, set $\gamma_{\textsc{a}} = 1$),
introducing \emph{bias} (see, e.g., the default PPO, Algorithm~\ref{algo:ppo} in the Appendix).
By adding this missing $\gamma_\textsc{a}^t$ term back
(\emph{i.e.}, setting $\gamma_\textsc{a} = \gamma < 1$),
we end up with an unbiased implementation, 
which we refer to as DisPPO (Algorithm~\ref{algo:dis-ppo} in the Appendix).
Figure~\ref{fig:aux_ppo},
however,
shows that even if we use the \emph{discounted} return as the performance metric,
the biased implementation of PPO still outperforms the theoretically grounded unbiased implementation DisPPO in some tasks.\footnote{In this scenario, by a task we mean the combination of a game and a discount factor.}
We propose to interpret the empirical advantages of PPO over DisPPO with Hypothesis~\ref{hyp:discounted}. 
For all experiments in this section,
we use $\gamma_{\textsc{c}} = \gamma < 1$.

\begin{figure}
\includegraphics[width=0.5\linewidth]{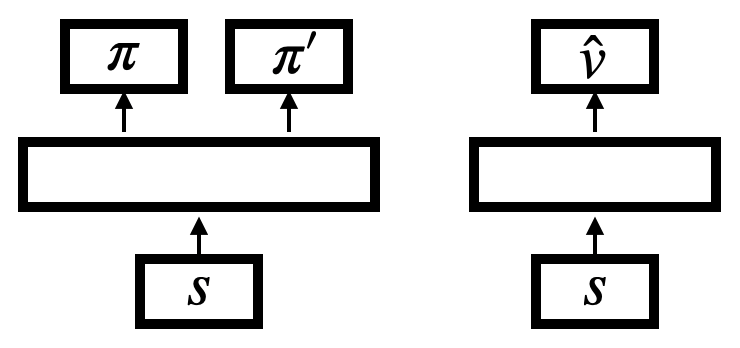}
\caption{\label{fig:auxppo arch} Architecture of AuxPPO}
\end{figure}

\textbf{An auxiliary task perspective: } The biased policy update implementation of \eqref{eq:pg_update} ignoring $\gamma_{\textsc{a}}^t$ can be decomposed into two parts as
\begin{align}
  &q_\pi^{\gamma_{\textsc{c}}}(S_t, A_t) \nabla_\theta \log \pi(A_t | S_t) \\
  =& \gamma^t q_\pi^{\gamma_{\textsc{c}}}(S_t, A_t) \nabla_\theta \log \pi(A_t | S_t) \\
  & + (1 - \gamma^t) q_\pi^{\gamma_{\textsc{c}}}(S_t, A_t) \nabla_\theta \log \pi(A_t | S_t). 
\end{align}
We propose to interpret the \emph{difference term} between the biased implementation $q_\pi^{\gamma_{\textsc{c}}}(S_t, A_t) \nabla_\theta \log \pi(A_t | S_t)$ and the theoretically grounded implementation $\gamma^t q_\pi^{\gamma_{\textsc{c}}}(S_t, A_t) \nabla_\theta \log \pi(A_t | S_t)$,
\textit{i.e.}, the $(1 - \gamma^t) q_\pi^{\gamma_{\textsc{c}}}(S_t, A_t) \nabla_\theta \log \pi(A_t | S_t)$ term,
as the gradient of an auxiliary objective with a dynamic weighting $1 - \gamma^t$.
Let 
\begin{align}
  \textstyle J_{s, \mu}(\pi) \doteq \sum_a \pi(a|s) q_\mu^\gamma(s, a),
\end{align}
we have 
\begin{align}
  \nabla_\theta J_{s, \mu}(\pi) |_{\mu = \pi} = \E_{a \sim \pi(\cdot | s)}[q_\pi^\gamma(s, a) \nabla_\theta \log \pi (a | s)].
\end{align}
This objective changes every time step (through $\mu$).
Inspired by the decomposition,
we augment PPO with this auxiliary task, yielding AuxPPO (Algorithm~\ref{algo:aux-ppo} in the appendix).
In AuxPPO, 
we have two policies $\pi$ and $\pi^\prime$
parameterized by $\theta$ and $\theta^\prime$ respectively.
The two policies are two heads over the same neural network backbone,
where $\pi$ is used for interaction with the environment and $\pi^\prime$ is the policy for the auxiliary task.
AuxPPO optimizes $\theta$ and $\theta^\prime$ simultaneously by considering the following joint loss
\begin{align*}
 L(\theta, &\theta^\prime) \doteq \textstyle\sum_{t=0}^\infty \gamma^t L_t(\theta, \theta_\text{old}) + (1 - \gamma^t) L_t(\theta', \theta_\text{old}),
\end{align*}
where $S_t$ and $A_t$ are obtained by executing $\pi_{\theta_\text{old}}$.
We additionally synchronize $\theta^\prime$ with $\theta$ periodically to avoid an off-policy learning issue. 
By contrast,
the objectives for PPO and DisPPO are
$\textstyle\sum_{t=0}^\infty \gamma^t L_t(\theta, \theta_\text{old}) + (1 - \gamma^t) L_t(\theta, \theta_\text{old})$ and $\textstyle\sum_{t=0}^\infty \gamma^t L_t(\theta, \theta_\text{old})$ respectively.
Figure~\ref{fig:auxppo arch} further illustrates the architecture of AuxPPO.

\textbf{Flipped rewards: }Besides AuxPPO,
we also design novel environments with flipped rewards to investigate Hypothesis~\ref{hyp:discounted}.
Recall we include the time step in the state,
which allows us to create a new environment by simply defining a new reward function 
\begin{align}
  r^\prime(s, t) \doteq r(s)\mathbb{I}_{t \leq t_0} - r(s) \mathbb{I}_{t > t_0},
\end{align}
where $\mathbb{I}$ is the indicator function.
During an episode,
within the first $t_0$ steps,
this new environment is the same as the original one.
After $t_0$ steps,
the sign of the reward is flipped.
We select $t_0$ such that $\gamma^{t_0}$ is sufficiently small,
\textit{e.g.}, we define $t_0 \doteq \min_t \{\gamma^t < 0.05\}$.
With this criterion for selecting $t_0$,
the later transitions (\textit{i.e.}, transitions after $t_0$ steps) have little influence on the evaluation objective, the discounted return.
Consequently,
the later transitions affect the overall learning process mainly through representation learning.
DisPPO rarely makes use of the later transitions due to the $\gamma_\textsc{a}^t$ term in the gradient update.
AuxPPO makes use of the later transitions only through representation learning (\emph{i.e.}, through the training of $\pi'$).
PPO exploits the later transitions for representation learning and the later transitions also affect the control policy of PPO directly.

\textbf{Results: }When we consider the original environments, 
Figure~\ref{fig:aux_ppo} shows that
in 8 out 12 tasks,
PPO outperforms DisPPO,
even if the performance metric is the \emph{discounted} episodic return.
In all those 8 tasks,
by using the difference term as an auxiliary task,
AuxPPO is able to improve upon DisPPO.
In 5 out of those 8 tasks,
AuxPPO is able to roughly match the performance of PPO at the end of training.
For $\gamma \in \{0.95, 0.93, 0.9\}$ in \texttt{Ant},
the improvement of AuxPPO is not clear and we conjecture that this is because the learning of the $\pi$-head (the control head) in AuxPPO is much slower than the learning of $\pi$ in PPO due to the $\gamma_{\textsc{c}}^t$ term.
Overall, this suggests that the benefit of PPO over DisPPO comes mainly from representation learning.

When we consider the environments with flipped rewards,
PPO is outperformed by DisPPO and AuxPPO by a large margin in 10 out of 12 tasks.
The transitions after $t_0$ steps are not directly relevant when the performance metric is the discounted return.
However,
learning on those transitions may still improve representation learning provided that those transitions are similar to the earlier transitions,
which is the case in the original environments.
PPO and AuxPPO, therefore, outperform DisPPO.
However,
when those transitions are much different from the earlier transitions,
which is the case in the environments with flipped rewards,
updating the control policy $\pi_\theta$ directly based on those transitions 
becomes distracting.
DisPPO, therefore, outperforms PPO.
Unlike PPO,
AuxPPO does not update the control policy $\pi_\theta$ on later transitions directly.
Provided that the network has enough capacity,
the irrelevant transitions do not affect the control policy $\pi_\theta$  in AuxPPO much.
The performance of AuxPPO is, therefore,
similar to that of DisPPO.

To summarize, Figure~\ref{fig:aux_ppo} suggests that using $\gamma_{\textsc{a}} = 1$ is simply an \emph{inductive bias} that \emph{all transitions are equally important}.
When this inductive bias is helpful for learning,
$\gamma_{\textsc{a}} = 1$ implicitly implements auxiliary tasks thus improving representation learning and the overall performance.
When this inductive bias is detrimental, however,
$\gamma_{\textsc{a}} = 1$ can lead to significant performance drops.
AuxPPO appears to be a safe choice that does not depend much on the correctness of this inductive bias.


\section{Related Work}
\label{sec related work}
The mismatch in actor-critic algorithm implementations has been previously studied.
\citet{thomas2014bias} focuses on the natural policy gradient setting and shows that the biased implementation ignoring $\gamma_{\textsc{a}}^t$ can be interpreted as the gradient of the average reward objective under a strong assumption that the state distribution is independent of the policy. 
\citet{DBLP:journals/corr/abs-1906-07073} prove that without this strong assumption, 
the biased implementation is \emph{not} the gradient of any \emph{stationary} objective. 
This does not contradict our auxiliary task perspective as our objective $J_{s, \mu}(\pi)$ changes at every time step.
\citet{DBLP:journals/corr/abs-1906-07073} further provide a counterexample showing that following the biased gradient can lead to a poorly performing policy w.r.t.\ both discounted and undiscounted objectives.
Both \citet{thomas2014bias} and \citet{DBLP:journals/corr/abs-1906-07073}, however,
focus on \emph{theoretical disadvantages} of the biased gradient and regard ignoring $\gamma_{\textsc{a}}^t$ as the source of the bias.
We instead regard the introduction of $\gamma_{\textsc{c}} < 1$ in the critic as the source of the bias in the undiscounted setting and investigate its \emph{empirical advantages},
which are more relevant to practitioners.
Moreover,
our representation learning perspective for investigating this mismatch is to our knowledge novel.
The concurrent work \citet{tang2021taylor} regards the biased implementation with $\gamma_{\textsc{a}} = 1$ as a partial gradient.
\citet{tang2021taylor},
however,
do not explain why this partial gradient can 
lead to empirical advantages over the full gradient.
The concurrent work \citet{laroche2021dr} proves that in the second scenario,
the biased setup can also converge to the optimal policy in the tabular setting,
assuming we have access to the transition kernel and the true value function.
Their results,
however,
heavily rely on the properties of the tabular setting and 
do not apply to the function approximation setting
we consider.

Although we propose the \emph{bias-variance-representation} trade-off,
we do not claim that is all that $\gamma$ affects.
The discount factor also has many other effects (\textit{e.g.}, \citet{sutton1995td,jiang2016structural,laroche2017multi,laroche2018reinforcement,lehnert2018value,fedus2019hyperbolic,van2019using,amit2020discount}),
the analysis of which we leave for future work.
In Scenario 1, using $\gamma_\textsc{c} < 1$ helps reduce the variance. 
Variance reduction in RL itself is an active research area (see, e.g., \citet{papini2018stochastic,xu2019sample,yuan2020stochastic}). 
Investigating those variance reduction techniques with $\gamma_\textsc{c} = 1$ is another possibility for future work.
Recently,
\citet{bengio2020interference} study the effect of the bootstrapping parameter $\lambda$ in TD($\lambda$) in generalization.
Our work studies the effect of the discount factor $\gamma$ in representation learning in the context of the misuse of the discounting in actor-critic algorithms,
sharing a similar spirit of \citet{bengio2020interference}. 


\section{Conclusion}
In this paper,
we investigated the longstanding mismatch between theory and practice in actor-critic algorithms from a representation learning perspective.
Although the theoretical understanding of policy gradient algorithms has recently advanced significantly~(\citet{agarwal2019optimality,wu2020finite}),
this mismatch has drawn little attention. 
We proposed to understand this mismatch from a bias-representation trade-off perspective and an auxiliary task perspective for two different scenarios.
We hope our empirical study can help practitioners understand actor-critic algorithms better and therefore design more efficient actor-critic algorithms in the setting of deep RL,
where representation learning emerges as a major consideration, as well as draw more attention to the mismatch,
which could enable the community to finally close this longstanding gap.

\begin{acks}
We thank Geoffrey J. Gordon, Marc-Alexandre Cote, Bei Peng, and Dipendra Misra for the insightful discussion. Part of this work was done during SZ's internship at Microsoft Research Montreal. SZ is also funded by the Engineering and Physical Sciences Research Council (EPSRC). This project has received funding from the European Research Council under the European Union's Horizon 2020 research and innovation programme (grant agreement number 637713). Part of the experiments was made possible by a generous equipment grant from NVIDIA.
\end{acks}



\bibliographystyle{ACM-Reference-Format} 
\bibliography{ref}


\newpage
\appendix
\onecolumn

\section{Proof of Lemma~\ref{lem:trpo-undiscounted}}
\begin{proof}
The proof is based on Appendix B in \citet{schulman2015trust},
where perturbation theory is used to prove the performance improvement bound (Lemma~\ref{lem:trpo}).
To simplify notation,
we use a vector and a function interchangeably,
\textit{i.e.},
we also use $r$ and $\mu_0$ to denote the reward vector and the initial distribution vector.
$J(\pi)$ and $d_\pi(s)$ are shorthand for $J_\gamma(\pi)$ and $d_\pi^\gamma(s)$ with $\gamma = 1$.
All vectors are \emph{column} vectors.

Let $\mathcal{S}^+$ be the set of states excluding $s^\infty$,
\textit{i.e.}, $\mathcal{S}^+ \doteq \mathcal{S} / \{s^\infty\}$,
we define $P_\pi \in \R^{\ns \times \ns}$ such that $P_\pi(s, s^\prime) \doteq \sum_a \pi(a|s)p(s^\prime | s, a)$.
Let $G \doteq \sum_{t=0}^\infty P_\pi^t$.
According to standard Markov chain theories, 
$G(s, s^\prime)$ is the expected number of times that $s^\prime$ is visited before $s^\infty$ is hit given $S_0 = s$. 
$T_{\max} < \infty$ implies that $G$ is well-defined and we have $G = (I - P_\pi)^{-1}$. 
Moreover, $T_{\max} < \infty$ also implies
$\forall s, \sum_{s^\prime} G(s, s^\prime) \leq T_{\max}$,
\textit{i.e.}, $||G||_\infty \leq T_{\max}$.
We have $J(\pi) = \mu_0^\top G r$.

Let $G^\prime \doteq (I - P_{\pi^\prime})^{-1}$,
we have 
\begin{align*}
J(\pi^\prime) - J(\pi) = \mu_0^\top(G^\prime - G)r.
\end{align*}
Let $\Delta \doteq P_{\pi^\prime} - P_\pi$,
we have 
\begin{align*}
G^{\prime-1} - G^{-1} = -\Delta,
\end{align*}
Left multiply by $G^\prime$ and right multiply by $G$,
\begin{align*}
G - G^\prime &= -G^\prime\Delta G, \\
G^\prime &= G + G^\prime \Delta G \quad \text{(Expanding $G^\prime$ in RHS recursively)} \\
 &= G + G \Delta G + G^\prime \Delta G \Delta G.
\end{align*}
So we have
\begin{align*}
J(\pi^\prime) - J(\pi) = \mu_0^\top G \Delta G r + \mu_0^\top G^\prime \Delta G \Delta G r.
\end{align*}
It is easy to see $\mu_0^\top G = d_\pi^\top$ and $Gr = v_\pi$. So
\begin{align*}
\mu_0^\top G \Delta G r &= d_\pi^\top \Delta v_\pi \\
&= \sum_s d_\pi(s) \sum_{s^\prime} \Big(\sum_a \pi^\prime(a|s) p(s^\prime|s, a) - \sum_a \pi(a|s) p(s^\prime|s, a)\Big) v_\pi(s^\prime) \\
&= \sum_s d_\pi(s) \sum_a (\pi^\prime(a |s) - \pi(a|s)) \sum_{s^\prime}p(s^\prime|s, a) v_\pi(s^\prime) \\
&= \sum_s d_\pi(s) \sum_a (\pi^\prime(a |s) - \pi(a|s)) \Big( r(s) + \sum_{s^\prime}p(s^\prime|s, a) v_\pi(s^\prime) - v_\pi(s) \Big) \\
\intertext{\hfill ($\sum_a(\pi^\prime(a | s) - \pi(a|s))f(s) = 0 $ holds for any $f$ that dependes only on $s$)}
&= \sum_s d_\pi(s) \sum_a \pi^\prime(a |s) \text{Adv}_\pi(s, a).
\intertext{\hfill ($\sum_a \pi(a|s) \text{Adv}_\pi(s, a)= 0$ by Bellman equation)}
\end{align*}
We now bound $\mu_0^\top G^\prime \Delta G \Delta G r$.
First,
\begin{align*}
|(\Delta G r)(s)| &= |\sum_{s^\prime} \Big( \sum_a \pi^\prime(a | s) - \pi(a|s) \Big) p(s^\prime | s, a) v_\pi(s^\prime)| \\
&= |\sum_a \Big(\pi^\prime(a|s) - \pi(a|s) \Big)\Big(r(s) + \sum_{s^\prime}p(s^\prime|s, a)v_\pi(s^\prime) - v_\pi(s) \Big)| \\
&= |\sum_a \Big(\pi^\prime(a|s) - \pi(a|s) \Big) \text{Adv}_\pi(s, a)| \\
&\leq 2\max_s\text{D}_{TV}(\pi^\prime(\cdot|s), \pi(\cdot|s) ) \max_{s,a}|\text{Adv}_\pi(s, a)|,
\end{align*}
where $\text{D}_{TV}$ is the total variation distance. So
\begin{align*}
||\Delta Gr||_\infty \leq 2\max_s\text{D}_{TV}(\pi^\prime(\cdot|s), \pi(\cdot|s) ) \max_{s,a}|\text{Adv}_\pi(s, a)|.
\end{align*}
Moreover, for any vector $x$,
\begin{align*}
|(\Delta x)(s)| &\leq 2\max_s\text{D}_{TV}(\pi^\prime(\cdot|s), \pi(\cdot|s) ) ||x||_\infty, \\
||\Delta x||_\infty &\leq 2\max_s\text{D}_{TV}(\pi^\prime(\cdot|s), \pi(\cdot|s) ) ||x||_\infty.
\end{align*}
So 
\begin{align*}
||\Delta||_\infty &\leq 2\max_s\text{D}_{TV}(\pi^\prime(\cdot|s), \pi(\cdot|s) ), \\
|\mu_0^\top G^\prime \Delta G \Delta G r| & \leq ||\mu_0^\top||_1 ||G^\prime||_\infty ||\Delta||_\infty ||G||_\infty ||\Delta Gr||_\infty \\
& \leq 4 T_{\max}^2 \max_s\text{D}^2_{TV}(\pi^\prime(\cdot|s), \pi(\cdot|s) ) \max_{s,a}|\text{Adv}_\pi(s, a)| \\
& \leq 4 T_{\max}^2 \max_s\text{D}_{KL}(\pi(\cdot|s) || \pi^\prime(\cdot|s) ) \max_{s,a}|\text{Adv}_\pi(s, a)|,
\end{align*}
which completes the proof.
\end{proof}
Note this perturbation-based proof of Lemma~\ref{lem:trpo-undiscounted} holds only for $r: \mathcal{S} \to \R$.
For $r: \mathcal{S} \times \mathcal{A} \to \R$,
we can turn to the coupling-based proof as \citet{schulman2015trust},
which, however, complicates the presentation and deviates from the main purpose of this paper.
We, therefore, leave it for future work.

\section{Experiment Details}
We conducted our experiments on an Nvidia DGX-1 with PyTorch,
though we do not use the GPUs there.
\subsection{Methodology}
\label{sec:app_method}
We use \texttt{HalfCheetah}, \texttt{Walker}, \texttt{Hopper}, \texttt{Ant}, \texttt{Humanoid}, and \texttt{HumanoidStandup} as our benchmarks.
We exclude other tasks as we find PPO plateaus quickly there.
The tasks we consider have a hard time limit of 1000. 
Following \citet{pardo2018time},
we add time step information into the state,
\textit{i.e.},
there is an additional scalar $t/{1000}$ in the observation vector.
Following \citet{spinningup2018},
we estimate the KL divergence between the current policy $\theta$ and the sampling policy $\theta_\text{old}$ when optimizing the loss \eqref{eq:ppo-loss}.
When the estimated KL divergence is greater than a threshold, 
we stop updating the actor and update only the critic with current data.
We use Adam (\citet{kingma2014adam}) as the optimizer and
perform grid search for the initial learning rates of Adam optimizers.
Let $\alpha_A$ and $\alpha_C \doteq \beta \alpha_A$ be the learning rates for the actor and critic respectively.
For each experiment unit (\emph{i.e.} an algorithmic configuration and a task, c.f. a curve in a figure), 
we tune $\alpha_A \in \{0.125, 0.25, 0.5, 1, 2\} \times 3\cdot 10^{-4}$ and $\beta \in \{1, 3\}$ with grid search with 3 independent runs maximizing the average return of the last 100 training episodes.
In particular,
$\alpha_A = 3 \cdot 10^{-4}$ and $\beta = 3$ is roughly the default learning rates for the PPO implementation in \citet{spinningup2018}.
Overall, we find after removing GAE, smaller learning rates are preferred.

In the discounted setting,
we consider only \texttt{Ant}, \texttt{HalfCheetah} and their variants.
For \texttt{Walker2d}, \texttt{Hopper}, and \texttt{Humanoid},
we find the average episode length of all algorithms are smaller than $t_0$,
\textit{i.e.},
the flipped reward rarely takes effects.
For \texttt{HumanoidStandup}, the scale of the reward is too large. 
To summarize, 
other four environments are not well-suited for the purpose of our empirical study.

\subsection{Algorithm Details}
The pseudocode of all implemented algorithms are provide in Algorithms~\ref{algo:ppo} - \ref{algo:aux-ppo}.
For hyperparameters that are not included in the grid search,
we use the same value as \citet{baselines,spinningup2018}.
In particular,
for the rollout length, we set $K = 2048$.
For the optimization epochs,
we set $K_{opt} = 320$.
For the minibatch size, we set $B = 64$.
For the maximum KL divergence, 
we set $KL_{target} = 0.01$.
We clip $\frac{\pi_\theta(a|s)}{\pi_{\theta_{old}}(a|s)}$ into $[-0.2, 0.2]$.

We use two-hidden-layer neural networks for function approximation.
Each hidden layer has 64 hidden units and a $\tanh$ activation function.
The output layer of the actor network has a $\tanh$ activation function and is interpreted as the mean of an isotropic Gaussian distribution,
whose standard derivation is a global state-independent variable as suggested by \citet{schulman2015trust}.

\begin{algorithm}
\SetAlgoLined
\KwIn{\\
$\theta, \psi$: parameters of $\pi, \hat{v}$\;
$\alpha_A, \alpha_C$: Initial learning rates of the Adam optimizers for $\theta, \psi$\;
$K, K_{opt}, B$: rollout length, number of optimization epochs, and minibatch size\;
${KL}_{target}$: maximum KL divergence threshold
}
\;
$S_0 \sim \mu_0$ \\
\While{True}{
Initialize a buffer $M$ \;
$\theta_{old} \gets \theta$ \;
\For{$i = 0, \dots, K - 1$}{
$A_i \sim \pi_{\theta_{old}}(\cdot|S_i)$ \;
Execute $A_i$, get $R_{i+1}, S_{i+1}$ \;
\eIf{$S_{i+1}$ is a terminal state}{
  $m_i \gets 0, S_{i+1} \sim \mu_0$
}{
  $m_i \gets 1$
}
}
$G_K \gets \hat{v}(S_K)$\;
\For{$i = K- 1, \dots, 0$}{
  $G_i \gets R_{i+1} + \gamma_{\textsc{c}} m_i G_{i+1}$ \;
  $\text{Adv}_i \gets R_{i+1} + \gamma_{\textsc{c}} m_i \hat{v}_\psi(S_{i+1}) - \hat{v}_\psi(S_i)$ \;
  Store $(S_i, A_i, G_i, \text{Adv}_i)$ in $M$ \;
}
Normalize \text{Adv}$_i$ in $M$ as $\text{Adv}_i \gets \frac{\text{Adv}_i - \text{mean}(\{\text{Adv}_i\})}{\text{std}(\{\text{Adv}_i\})}$ \;
\For{$o = 1, \dots, K_{opt}$}{
  Sample a minibatch $\{(S_i, A_i, G_i, \text{Adv}_i)\}_{i = 1, \dots, B}$ from $M$ \;
  $L(\psi) \gets \frac{1}{2B}\sum_{i=1}^{B}(\hat{v}_\psi(S_i) - G_i)^2$ \tcc{No gradient through $G_i$}
  $L(\theta) \gets \frac{1}{B} \sum_{i=1}^B \min\{ \frac{\pi_\theta(A_i|S_i)}{\pi_{\theta_{old}}(A_i | S_i)}\text{Adv}_i, \text{clip}(\frac{\pi_\theta(A_i|S_i)}{\pi_{\theta_{old}}(A_i | S_i)})\text{Adv}_i\}$ \;
  Perform one gradient update to $\psi$ minimizing $L(\psi)$ with Adam\;
  \If{$\frac{1}{B}\sum_{i=1}^B \log \pi_{\theta_{old}}(A_i|S_i) - \log \pi_\theta(A_i|S_i) < KL_{target}$}{
  Perform one gradient update to $\theta$ maximizing $L(\theta)$ with Adam\;
  }
}
}
\caption{\label{algo:ppo}PPO}
\end{algorithm}

\begin{algorithm}
\SetAlgoLined
\KwIn{\\
$\theta, \psi$: parameters of $\pi, \hat{v}$\;
$\alpha_A, \alpha_C$: Initial learning rates of the Adam optimizers for $\theta, \psi$\;
$K, K_{opt}, B$: rollout length, number of optimization epochs, and minibatch size\;
${KL}_{target}$: maximum KL divergence threshold
}
\;
$S_0 \sim \mu_0$ \\
\While{True}{
Initialize a buffer $M$ \;
$\theta_{old} \gets \theta$ \;
\For{$i = 0, \dots, K - 1$}{
$A_i \sim \pi_{\theta_{old}}(\cdot|S_i)$ \;
Execute $A_i$, get $R_{i+1}, S_{i+1}$ \;
\eIf{$S_{i+1}$ is a terminal state}{
  $m_i \gets 0, S_{i+1} \sim \mu_0$
}{
  $m_i \gets 1$
}
}
\For{$i = K- 1, \dots, 0$}{
  $\text{Adv}_i \gets R_{i+1} + \gamma_{\textsc{c}} m_i \hat{v}_\psi(S_{i+1}) - \hat{v}_\psi(S_i)$ \; 
  $S_i^\prime \gets S_{i+1}, r_i \gets R_{i+1}$ \;
  Store $(S_i, A_i, m_i, r_i, S^\prime_i, \text{Adv}_i)$ in $M$\;
}
Normalize \text{Adv}$_i$ in $M$ as $\text{Adv}_i \gets \frac{\text{Adv}_i - \text{mean}(\{\text{Adv}_i\})}{\text{std}(\{\text{Adv}_i\})}$ \;
\For{$o = 1, \dots, K_{opt}$}{
  Sample a minibatch $\{(S_i, A_i, m_i, r_i, S_i^\prime, \text{Adv}_i)\}_{i = 1, \dots, B}$ from $M$ \;
  {\color{red} $y_i \gets r_i + \gamma_{\textsc{c}} m_i \hat{v}_\psi(S_i^\prime$)} \;
  $L(\psi) \gets \frac{1}{2B}\sum_{i=1}^{B}(\hat{v}_\psi(S_i) - y_i)^2$ \tcc{No gradient through $y_i$}
  $L(\theta) \gets \frac{1}{B} \sum_{i=1}^B \min\{ \frac{\pi_\theta(A_i|S_i)}{\pi_{\theta_{old}}(A_i | S_i)}\text{Adv}_i, \text{clip}(\frac{\pi_\theta(A_i|S_i)}{\pi_{\theta_{old}}(A_i | S_i)})\text{Adv}_i\}$ \;
  Perform one gradient update to $\psi$ minimizing $L(\psi)$ with Adam\;
  \If{$\frac{1}{B}\sum_{i=1}^B \log \pi_{\theta_{old}}(A_i|S_i) - \log \pi_\theta(A_i|S_i) < KL_{target}$}{
  Perform one gradient update to $\theta$ maximizing $L(\theta)$ with Adam\;
  }
}
}
\caption{\label{algo:ppo-td}PPO-TD}
\end{algorithm}

\begin{algorithm}
\SetAlgoLined
\KwIn{\\
$\theta, \psi$: parameters of $\pi, \hat{v}$\;
$\alpha_A, \alpha_C$: Initial learning rates of the Adam optimizers for $\theta, \psi$\;
$K, K_{opt}, B$: rollout length, number of optimization epochs, and minibatch size\;
${KL}_{target}$: maximum KL divergence threshold \;
$N$: number of extra transitions \;
$p, r$: transition kernel and reward function of the oracle 
}
\;
$S_0 \sim \mu_0$ \\
\While{True}{
Initialize a buffer $M$ \;
$\theta_{old} \gets \theta$ \;
\For{$i = 0, \dots, K - 1$}{
\For{$j = 0, \dots, N$}{
  $A_i^j \sim \pi_{\theta_{old}}(\cdot | S_i), R_{i+1}^j \gets r(S_i, A_i^j), S_{i+1}^j \sim p(\cdot|S_i, A_i^j)$ \;
  \eIf{$S_{i+1}^j$ is a terminal state}{
  $m_i^j \gets 0, S_{i+1}^j \sim \mu_0$
}{
  $m_i^j \gets 1$
}
}
$S_{i+1} \gets S_{i+1}^0$
}
\For{$i = K- 1, \dots, 0$}{
  $\text{Adv}_i \gets R_{i+1}^0 + m_i^0 \hat{v}_\psi(S_{i+1}^0) - \hat{v}_\psi(S_i^0)$ \; 
  \For{$j = 0, \dots, N$}{
    $S_i^{\prime j} \gets S_{i+1}^j$
  }
  Store $(\{S_i^j, A_i^j, m_i^j, r_i^j, S_i^{\prime j}\}_{j = 0, \dots, N}, \text{Adv}_i)$ in $M$\;
}
Normalize \text{Adv}$_i$ in $M$ as $\text{Adv}_i \gets \frac{\text{Adv}_i - \text{mean}(\{\text{Adv}_i\})}{\text{std}(\{\text{Adv}_i\})}$ \;
\For{$o = 1, \dots, K_{opt}$}{
  Sample a minibatch $\{(\{S_i^j, A_i^j, m_i^j, r_i^j, S_i^{\prime j}\}_{j = 0, \dots, N}, \text{Adv}_i)\}_{i = 1, \dots, B}$ from $M$ \;
  {\color{red} $y_i \gets \frac{1}{N+1} \sum_{j = 0}^N r_i^j + \gamma_{\textsc{c}} m_i^j \hat{v}_\psi(S_i^{\prime j}$)} \;
  $L(\psi) \gets \frac{1}{2B}\sum_{i=1}^{B}(\hat{v}_\psi(S_i^0) - y_i)^2$ \tcc{No gradient through $y_i$}
  $L(\theta) \gets \frac{1}{B} \sum_{i=1}^B \min\{ \frac{\pi_\theta(A_i^0|S_i^0)}{\pi_{\theta_{old}}(A_i^0 | S_i^0)}\text{Adv}_i, \text{clip}(\frac{\pi_\theta(A_i^0|S_i^0)}{\pi_{\theta_{old}}(A_i^0 | S_i^0)})\text{Adv}_i\}$ \;
  Perform one gradient update to $\psi$ minimizing $L(\psi)$ with Adam\;
  \If{$\frac{1}{B}\sum_{i=1}^B \log \pi_{\theta_{old}}(A_i^0|S_i^0) - \log \pi_\theta(A_i^0|S_i^0) < KL_{target}$}{
  Perform one gradient update to $\theta$ maximizing $L(\theta)$ with Adam\;
  }
}
}
\caption{\label{algo:ppo-td-ex}PPO-TD-Ex}
\end{algorithm}

\begin{algorithm}
\SetAlgoLined
\KwIn{\\
$\theta, \psi$: parameters of $\pi, \{\hat{v}^j\}_{j = 1, \dots, H}$\;
$\alpha_A, \alpha_C$: Initial learning rates of the Adam optimizers for $\theta, \psi$\;
$K, K_{opt}, B$: rollout length, number of optimization epochs, and minibatch size\;
${KL}_{target}$: maximum KL divergence threshold
}
\;
$S_0 \sim \mu_0$ \\
\While{True}{
Initialize a buffer $M$ \;
$\theta_{old} \gets \theta$ \;
\For{$i = 0, \dots, K - 1$}{
$A_i \sim \pi_{\theta_{old}}(\cdot|S_i)$ \;
Execute $A_i$, get $R_{i+1}, S_{i+1}$ \;
\eIf{$S_{i+1}$ is a terminal state}{
  $m_i \gets 0, S_{i+1} \sim \mu_0$
}{
  $m_i \gets 1$
}
}
\For{$i = K- 1, \dots, 0$}{
  $\text{Adv}_i \gets R_{i+1} + m_i \hat{v}_\psi^H(S_{i+1}) - \hat{v}_\psi^H(S_i)$ \; 
  $S_i^\prime \gets S_{i+1}, r_i \gets R_{i+1}$ \;
  Store $(S_i, A_i, m_i, r_i, S^\prime_i, \text{Adv}_i)$ in $M$\;
}
Normalize \text{Adv}$_i$ in $M$ as $\text{Adv}_i \gets \frac{\text{Adv}_i - \text{mean}(\{\text{Adv}_i\})}{\text{std}(\{\text{Adv}_i\})}$ \;
\For{$o = 1, \dots, K_{opt}$}{
  Sample a minibatch $\{(S_i, A_i, m_i, r_i, S_i^\prime, \text{Adv}_i)\}_{i = 1, \dots, B}$ from $M$ \;
  \For{$j = 1, \dots, H$}{
    $y_i^j \gets r_i + m_i \hat{v}_\psi^{j-1}(S_i^\prime))$ \quad \tcc{$\hat{v}^0(S_i^\prime) \equiv 0$}
  }
  $L(\psi) \gets \frac{1}{2B}\sum_{i=1}^{B} \sum_{j=1}^H (\hat{v}_\psi^j(S_i) - y_i^j)^2$ \tcc{No gradient through $y_i^j$}
  $L(\theta) \gets \frac{1}{B} \sum_{i=1}^B \min\{ \frac{\pi_\theta(A_i|S_i)}{\pi_{\theta_{old}}(A_i | S_i)}\text{Adv}_i, \text{clip}(\frac{\pi_\theta(A_i|S_i)}{\pi_{\theta_{old}}(A_i | S_i)})\text{Adv}_i\}$ \;
  Perform one gradient update to $\psi$ minimizing $L(\psi)$ with Adam\;
  \If{$\frac{1}{B}\sum_{i=1}^B \log \pi_{\theta_{old}}(A_i|S_i) - \log \pi_\theta(A_i|S_i) < KL_{target}$}{
  Perform one gradient update to $\theta$ maximizing $L(\theta)$ with Adam\;
  }
}
}
\caption{\label{algo:ppo-fhtd}PPO-FHTD}
\end{algorithm}

\begin{algorithm}
\SetAlgoLined
\KwIn{\\
$\theta, \psi$: parameters of $\pi, \hat{v}$\;
$\alpha_A, \alpha_C$: Initial learning rates of the Adam optimizers for $\theta, \psi$\;
$K, K_{opt}, B$: rollout length, number of optimization epochs, and minibatch size\;
${KL}_{target}$: maximum KL divergence threshold
}
\;
$S_0 \sim \mu_0, t \gets 0$ \\
\While{True}{
Initialize a buffer $M$ \;
$\theta_{old} \gets \theta$ \;
\For{$i = 0, \dots, K - 1$}{
$A_i \sim \pi_{\theta_{old}}(\cdot|S_i), t_i \gets t$ \;
Execute $A_i$, get $R_{i+1}, S_{i+1}$ \;
\eIf{$S_{i+1}$ is a terminal state}{
  $m_i \gets 0, S_{i+1} \sim \mu_0, t \gets 0$
}{
  $m_i \gets 1, t \gets t + 1$
}
}
$G_K \gets \hat{v}(S_K)$\;
\For{$i = K- 1, \dots, 0$}{
  $G_i \gets R_{i+1} + \gamma_{\textsc{c}} m_i G_{i+1}$ \;
  $\text{Adv}_i \gets R_{i+1} + \gamma_{\textsc{c}} m_i \hat{v}_\psi(S_{i+1}) - \hat{v}_\psi(S_i)$ \;
  Store $(S_i, A_i, G_i, \text{Adv}_i, t_i)$ in $M$ \;
}
Normalize \text{Adv}$_i$ in $M$ as $\text{Adv}_i \gets \frac{\text{Adv}_i - \text{mean}(\{\text{Adv}_i\})}{\text{std}(\{\text{Adv}_i\})}$ \;
\For{$o = 1, \dots, K_{opt}$}{
  Sample a minibatch $\{(S_i, A_i, G_i, \text{Adv}_i, t_i)\}_{i = 1, \dots, B}$ from $M$ \;
  $L(\psi) \gets \frac{1}{2B}\sum_{i=1}^{B}(\hat{v}_\psi(S_i) - G_i)^2$ \tcc{No gradient through $G_i$}
  $L(\theta) \gets \frac{1}{B} \sum_{i=1}^B {\color{red} \gamma_\textsc{A}^{t_i}} \min\{ \frac{\pi_\theta(A_i|S_i)}{\pi_{\theta_{old}}(A_i | S_i)}\text{Adv}_i, \text{clip}(\frac{\pi_\theta(A_i|S_i)}{\pi_{\theta_{old}}(A_i | S_i)})\text{Adv}_i\}$ \;
  Perform one gradient update to $\psi$ minimizing $L(\psi)$ with Adam\;
  \If{$\frac{1}{B}\sum_{i=1}^B \log \pi_{\theta_{old}}(A_i|S_i) - \log \pi_\theta(A_i|S_i) < KL_{target}$}{
  Perform one gradient update to $\theta$ maximizing $L(\theta)$ with Adam\;
  }
}
}
\caption{\label{algo:dis-ppo}DisPPO}
\end{algorithm}

\begin{algorithm}
\SetAlgoLined
\KwIn{\\
$\theta, \theta^\prime, \psi$: parameters of $\pi, \pi^\prime, \hat{v}$\;
$\alpha_A, \alpha_C$: Initial learning rates of the Adam optimizers for $\theta, \psi$\;
$K, K_{opt}, B$: rollout length, number of optimization epochs, and minibatch size\;
${KL}_{target}$: maximum KL divergence threshold
}
\;
$S_0 \sim \mu_0, t \gets 0$ \\
\While{True}{
Initialize a buffer $M$ \;
$\theta_{old} \gets \theta, {\color{red} \theta^\prime \gets \theta}$ \;
\For{$i = 0, \dots, K - 1$}{
$A_i \sim \pi_{\theta_{old}}(\cdot|S_i), t_i \gets t$ \;
Execute $A_i$, get $R_{i+1}, S_{i+1}$ \;
\eIf{$S_{i+1}$ is a terminal state}{
  $m_i \gets 0, S_{i+1} \sim \mu_0, t \gets 0$
}{
  $m_i \gets 1, t \gets t + 1$
}
}
$G_K \gets \hat{v}(S_K)$\;
\For{$i = K- 1, \dots, 0$}{
  $G_i \gets R_{i+1} + \gamma_{\textsc{c}} m_i G_{i+1}$ \;
  $\text{Adv}_i \gets R_{i+1} + \gamma_{\textsc{c}} m_i \hat{v}_\psi(S_{i+1}) - \hat{v}_\psi(S_i)$ \;
  Store $(S_i, A_i, G_i, \text{Adv}_i, t_i)$ in $M$ \;
}
Normalize \text{Adv}$_i$ in $M$ as $\text{Adv}_i \gets \frac{\text{Adv}_i - \text{mean}(\{\text{Adv}_i\})}{\text{std}(\{\text{Adv}_i\})}$ \;
\For{$o = 1, \dots, K_{opt}$}{
  Sample a minibatch $\{(S_i, A_i, G_i, \text{Adv}_i, t_i)\}_{i = 1, \dots, B}$ from $M$ \;
  $L(\psi) \gets \frac{1}{2B}\sum_{i=1}^{B}(\hat{v}_\psi(S_i) - G_i)^2$ \tcc{No gradient through $G_i$}
  \begin{align*}
  \textstyle{ L(\theta, \theta^\prime) \gets \frac{1}{B} \sum_{i=1}^B} & \textstyle{ {\color{red} \gamma_\textsc{C}^{t_i}} \min\{ \frac{\pi_\theta(A_i|S_i)}{\pi_{\theta_{old}}(A_i | S_i)}\text{Adv}_i, \text{clip}(\frac{\pi_\theta(A_i|S_i)}{\pi_{\theta_{old}}(A_i | S_i)})\text{Adv}_i\} + }\\
  &\textstyle{ {\color{red} (1 - \gamma_\textsc{C}^{t_i})} \min\{ \frac{\pi_{\color{red} \theta^\prime}(A_i|S_i)}{\pi_{\theta_{old}}(A_i | S_i)}\text{Adv}_i, \text{clip}(\frac{\pi_{\color{red} \theta^\prime}(A_i|S_i)}{\pi_{\theta_{old}}(A_i | S_i)})\text{Adv}_i\} }
  \end{align*} \;
  Perform one gradient update to $\psi$ minimizing $L(\psi)$ with Adam\;
  \If{$\frac{1}{B}\sum_{i=1}^B \log \pi_{\theta_{old}}(A_i|S_i) - \log \pi_\theta(A_i|S_i) < KL_{target}$}{
  Perform one gradient update to $\theta, \theta^\prime$ maximizing $L(\theta, \theta^\prime)$ with Adam\;
  }
}
}
\caption{\label{algo:aux-ppo}AuxPPO}
\end{algorithm}

\newpage
\section{Additional Experimental Results}

\begin{figure}[h]
\centering
\includegraphics[width=\linewidth]{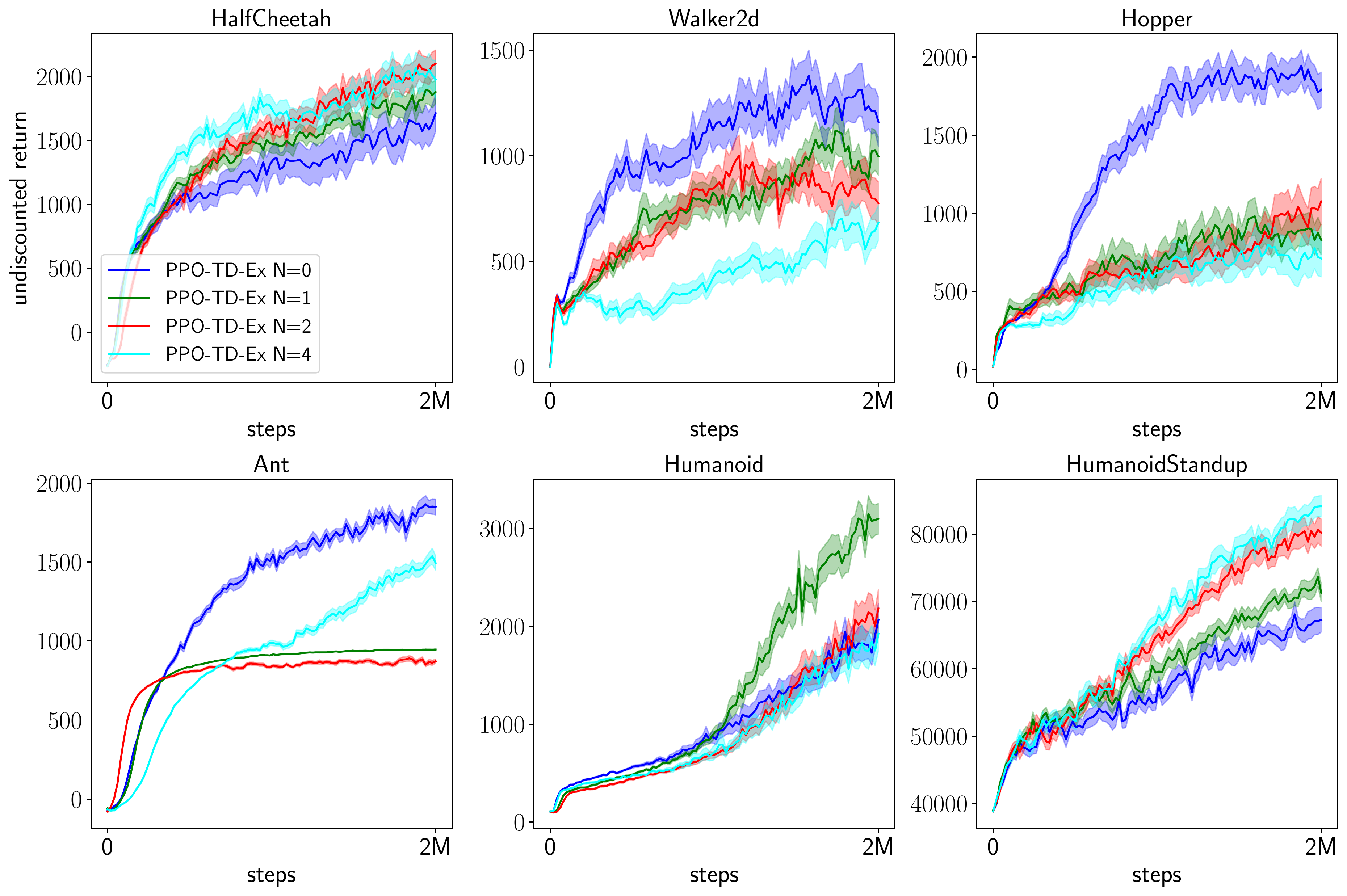}
\caption{\label{fig:ppo-td-mp-ret-0.995} PPO-TD-Ex ($\gamma_{\textsc{c}} = 0.995$).}
\end{figure}

\begin{figure}[h]
\centering
\includegraphics[width=0.9\linewidth]{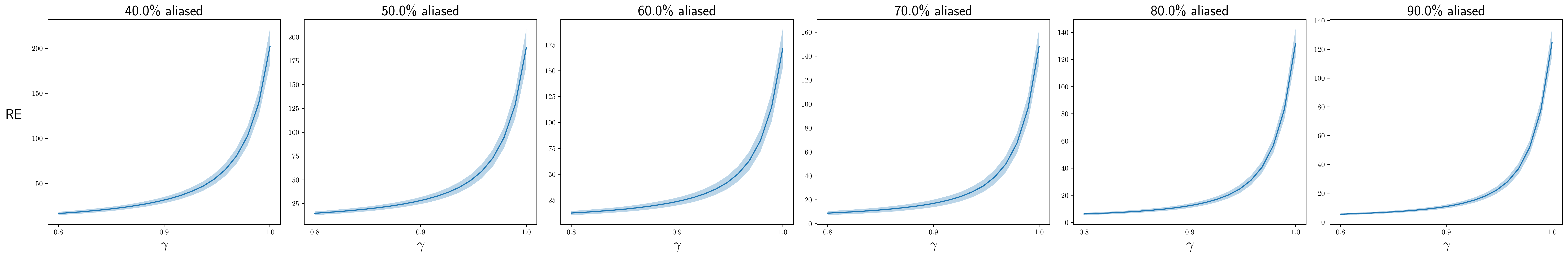}
\caption{\label{fig:state_aliasing_false} Unnormalized representation error (RE) as a function of the discount factor.
Shaded regions indicate one standard derivation.
RE is computed analytically as \\$\text{RE}(X, \gamma) \doteq \min_w ||Xw - v_\gamma||_2$}
\end{figure}

\begin{figure}[h]
\centering
\includegraphics[width=\linewidth]{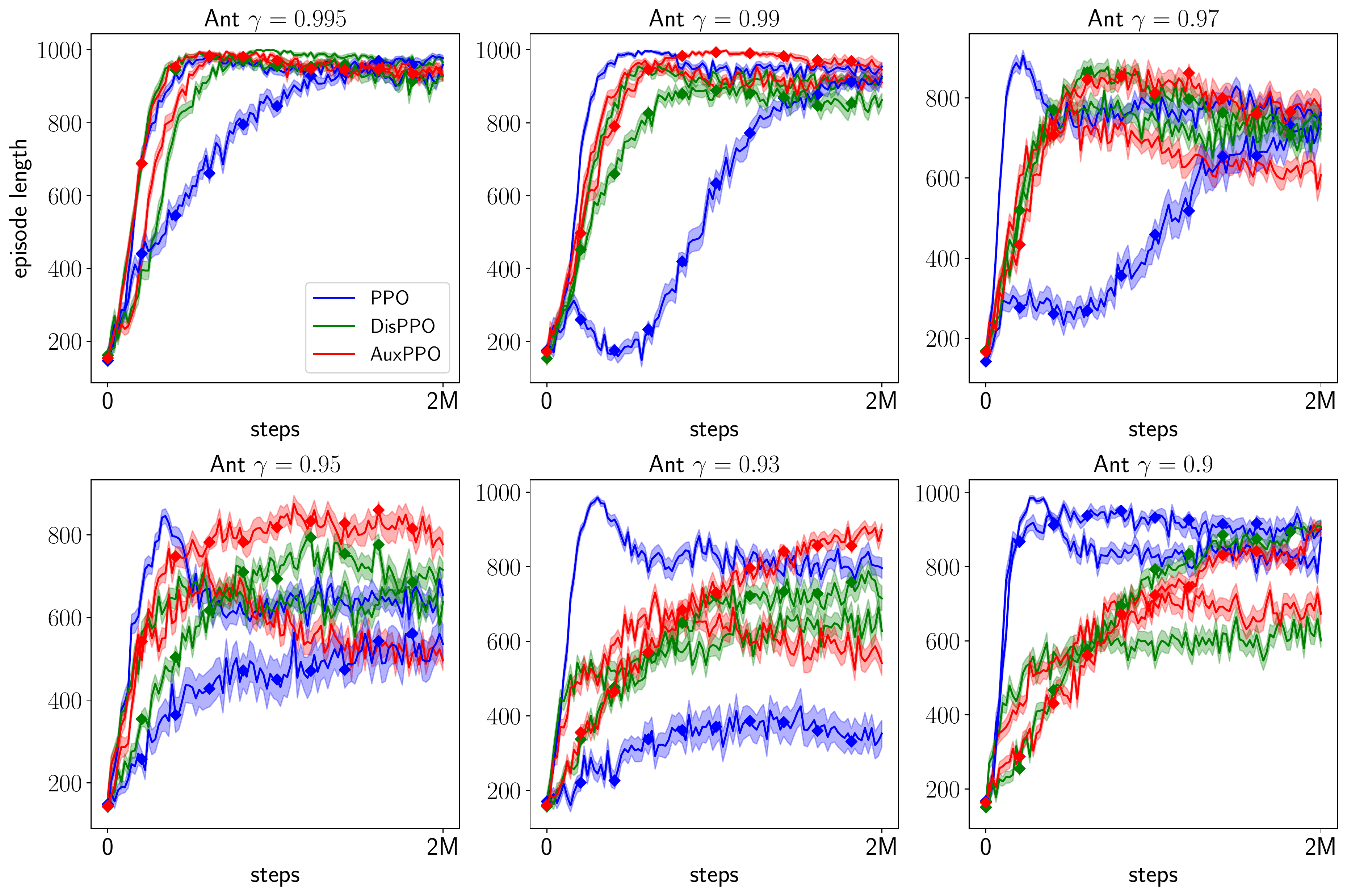}
\caption{\label{fig:aux-ppo-episode-len} 
Curves without any marker are obtained in the original \texttt{Ant}.
Diamond-marked curves are obtained in \texttt{Ant} with $r^\prime$.}
\end{figure}

Figure~\ref{fig:ppo-td-mp-ret-0.995} shows how PPO-TD-Ex ($\gamma_\textsc{c} = 0.995$) reacts to the increase of $N$.
Figure~\ref{fig:state_aliasing_false} shows the unnormalized representation error in the MRP experiment.
Figure~\ref{fig:aux-ppo-episode-len} shows the average episode length for the \texttt{Ant} environment in the discounted setting.
For \texttt{HalfCheetah}, it is always 1000.


\section{Larger Version of Figures}

\begin{figure}[h]
\centering
\includegraphics[width=\linewidth]{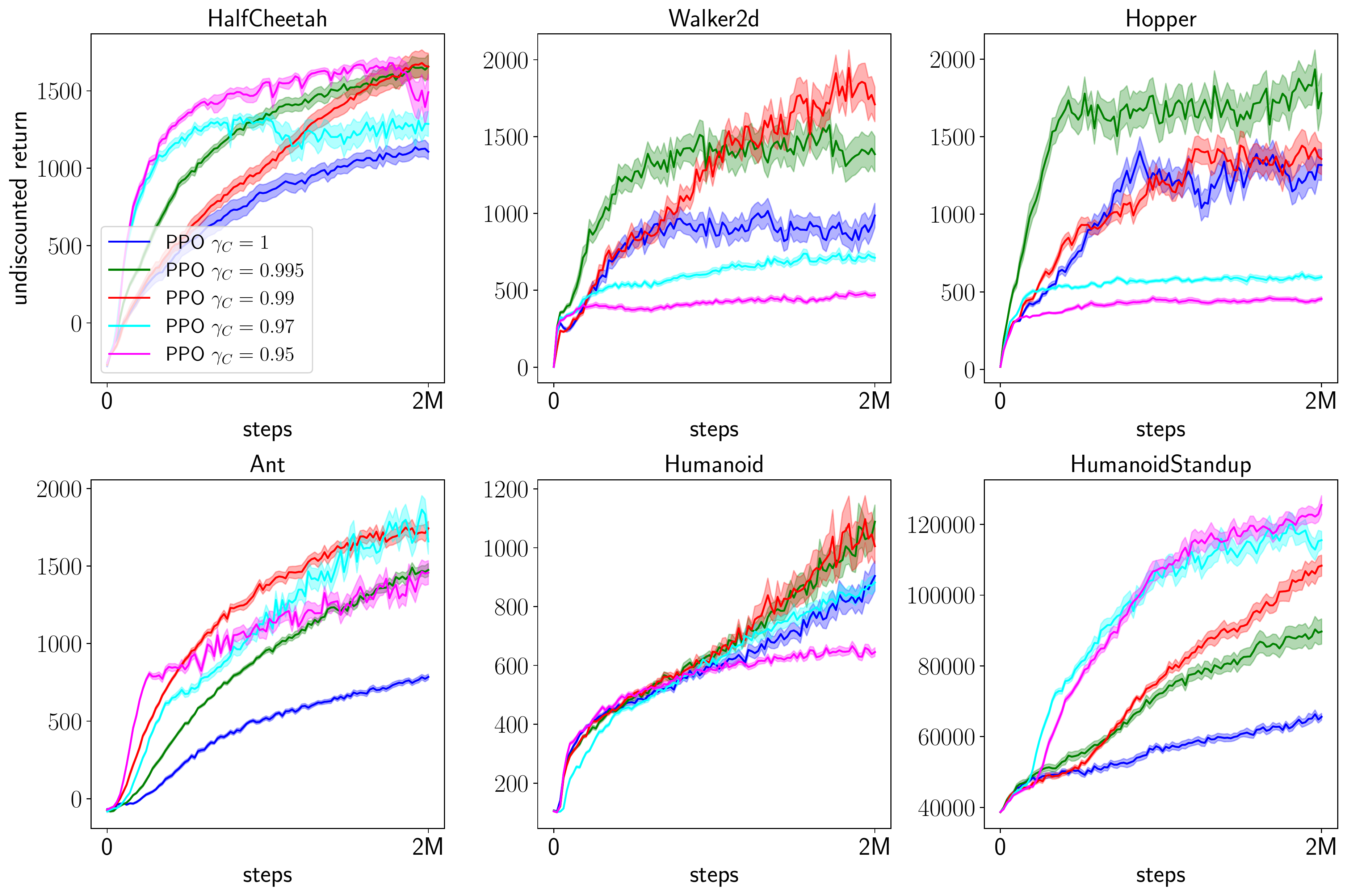}
\caption{The default PPO implementation with different discount factors. The larger version of Figure~\ref{fig:ppo-ret}.}
\end{figure}
\begin{figure}[h]
\centering
\includegraphics[width=\linewidth]{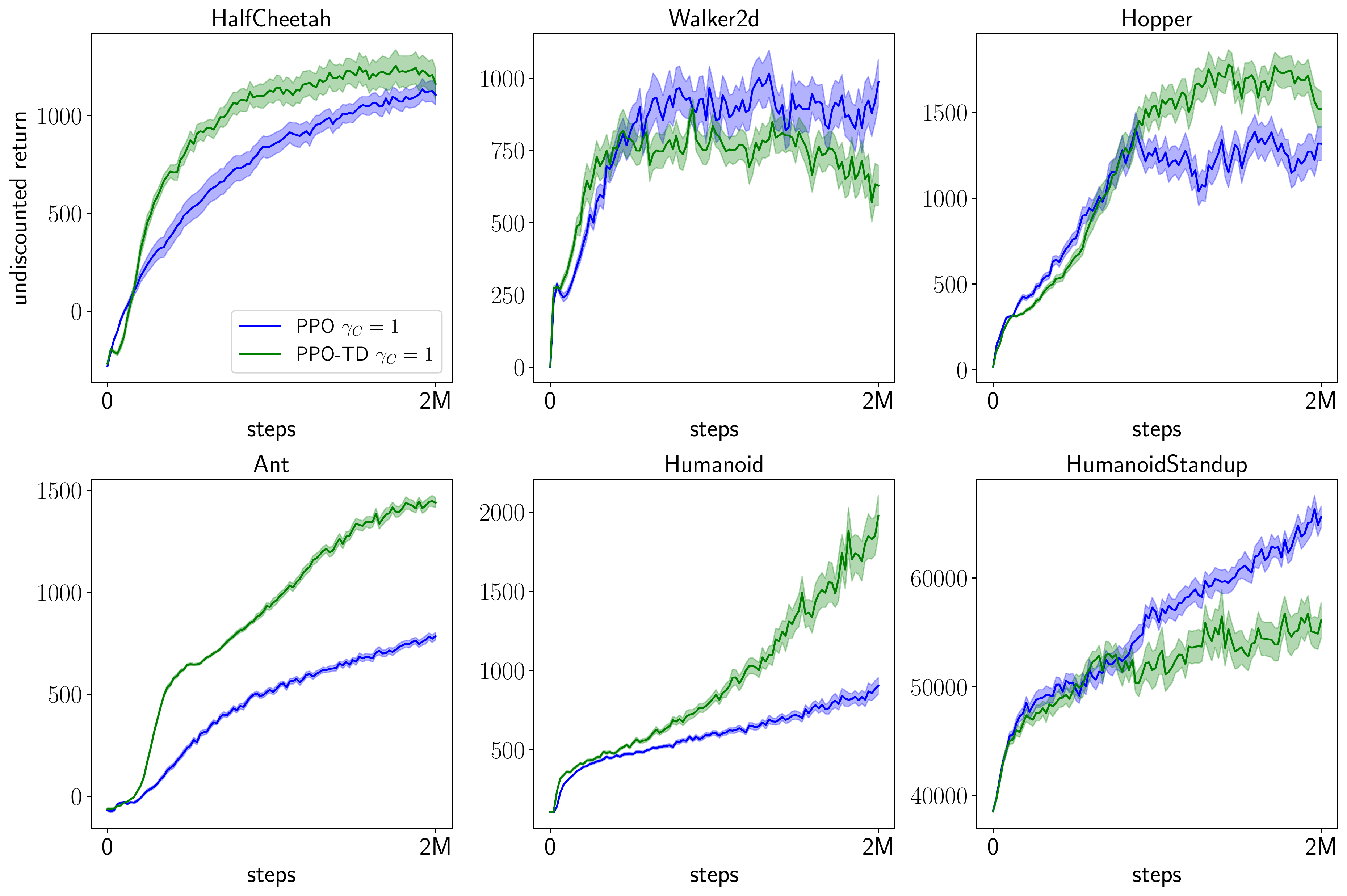}
\caption{Comparison between PPO and PPO-TD when $\gamma_{\textsc{c}} = 1$. The larger version of Figure~\ref{fig:ppo-td-mc-ret}.}
\end{figure}
\begin{figure}[h]
\centering
\includegraphics[width=\linewidth]{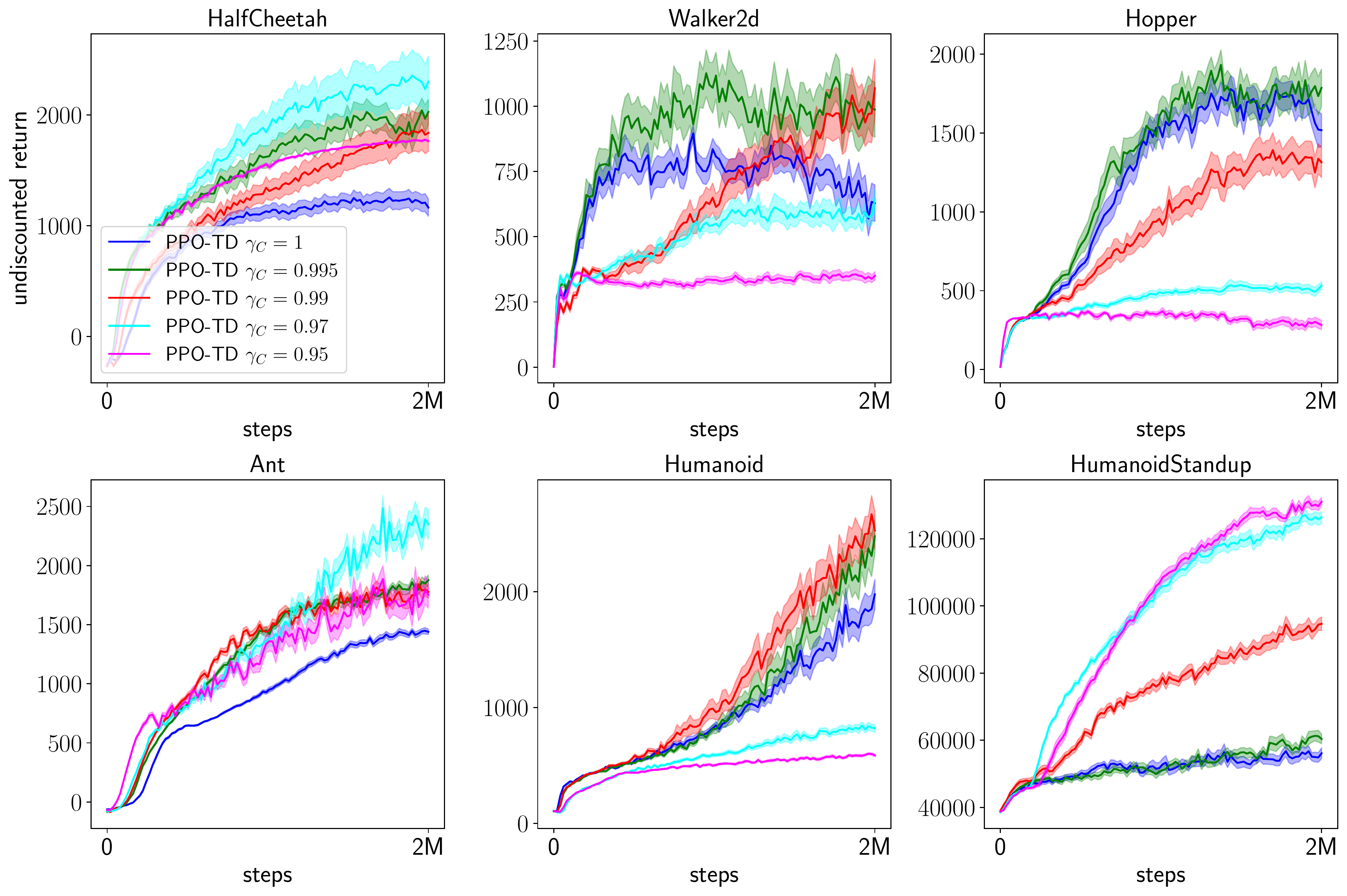}
\caption{PPO-TD with different discount factors. The larger version of Figure~\ref{fig:ppo-td-critic-ret}.}
\end{figure}
\begin{figure}[h]
\centering
\includegraphics[width=\linewidth]{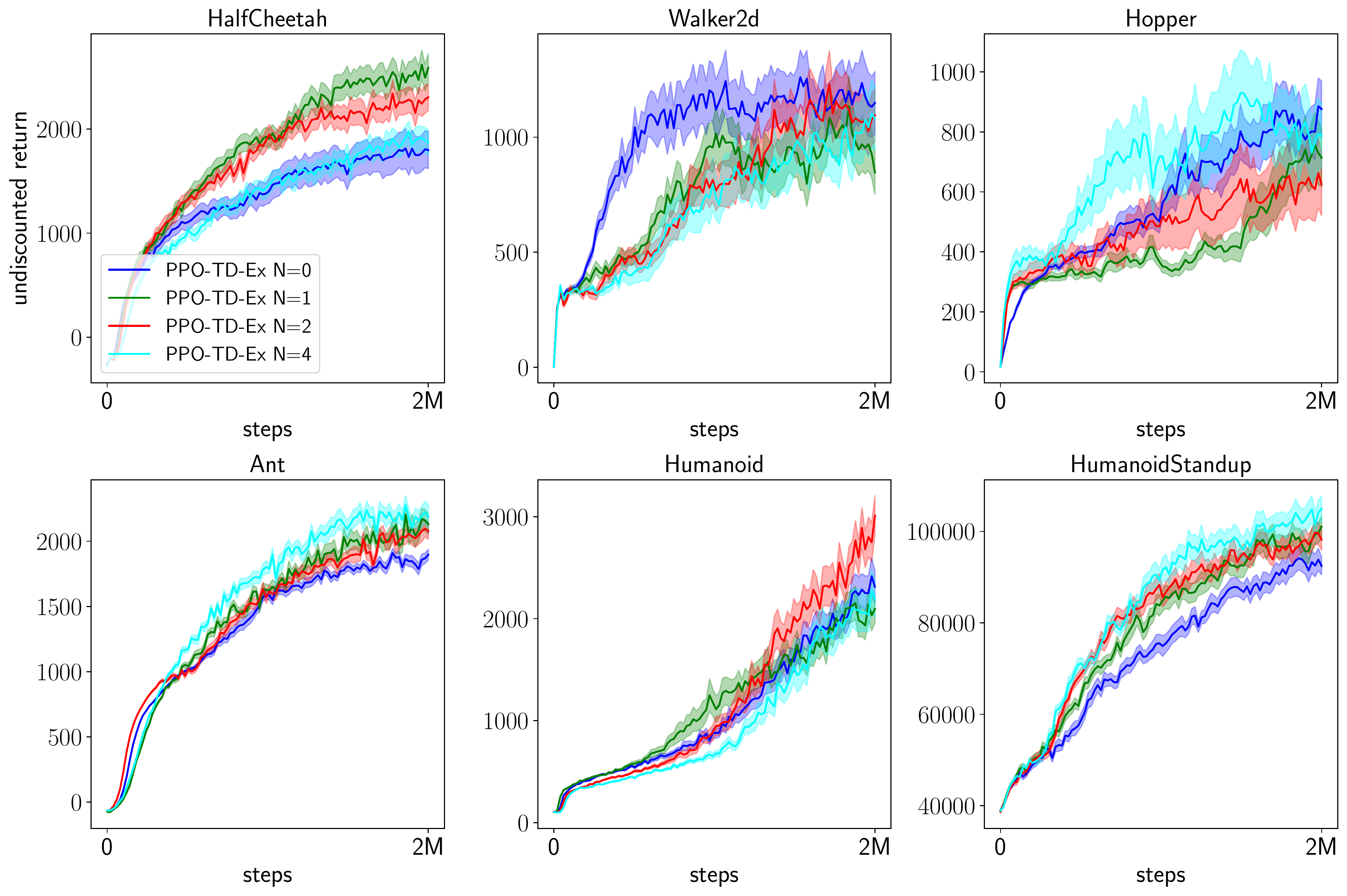}
\caption{PPO-TD-Ex ($\gamma_{\textsc{c}} = 0.99$). The larger version of Figure~\ref{fig:ppo-td-mp-ret-0.99}.}
\end{figure}
\begin{figure}[h]
\centering
\includegraphics[width=\linewidth]{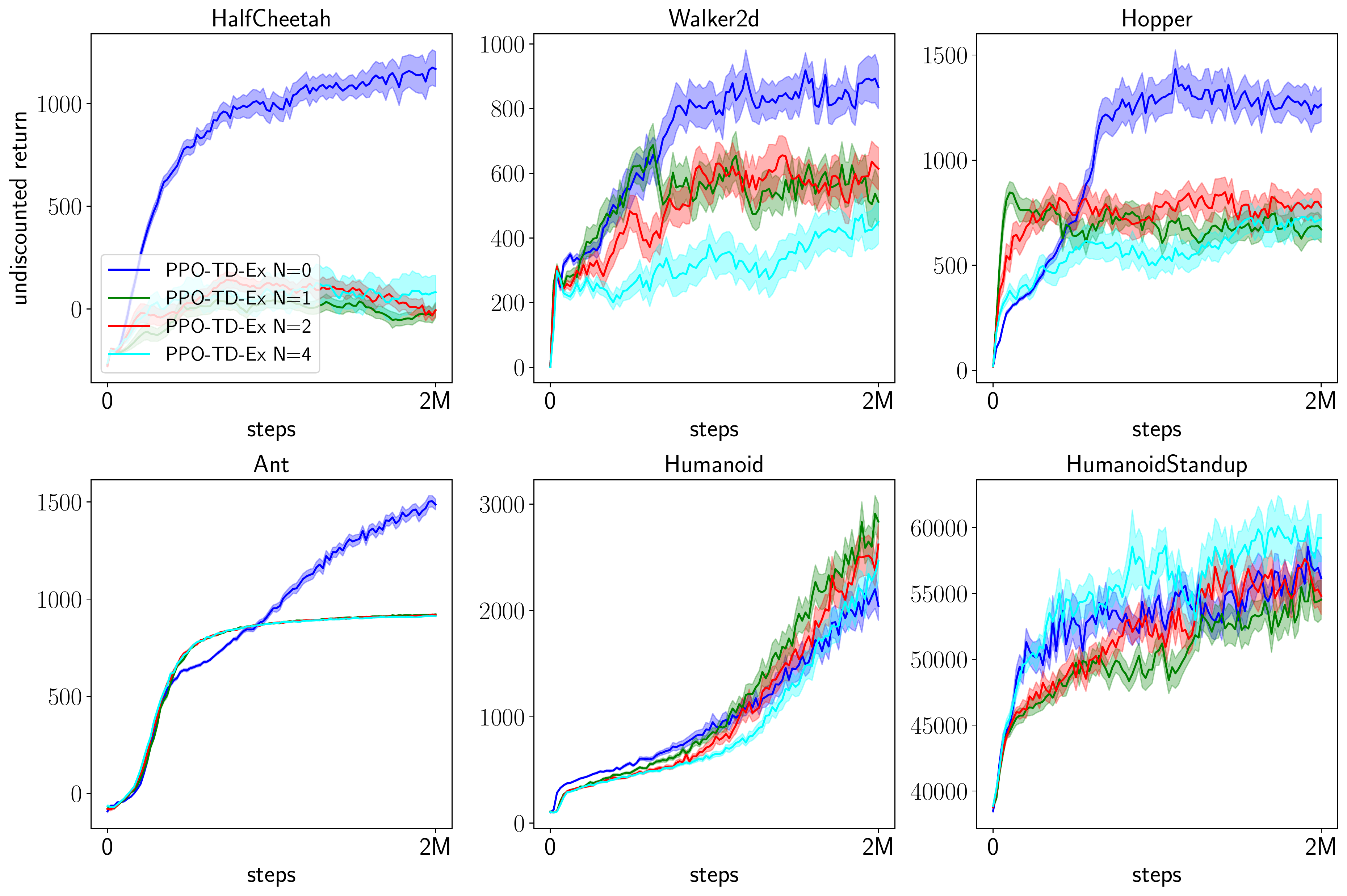}
\caption{PPO-TD-Ex ($\gamma_{\textsc{c}} = 1$). The larger version of Figure~\ref{fig:ppo-td-mp-ret-1}.}
\end{figure}

\begin{figure}[h]
\centering
\includegraphics[width=\linewidth]{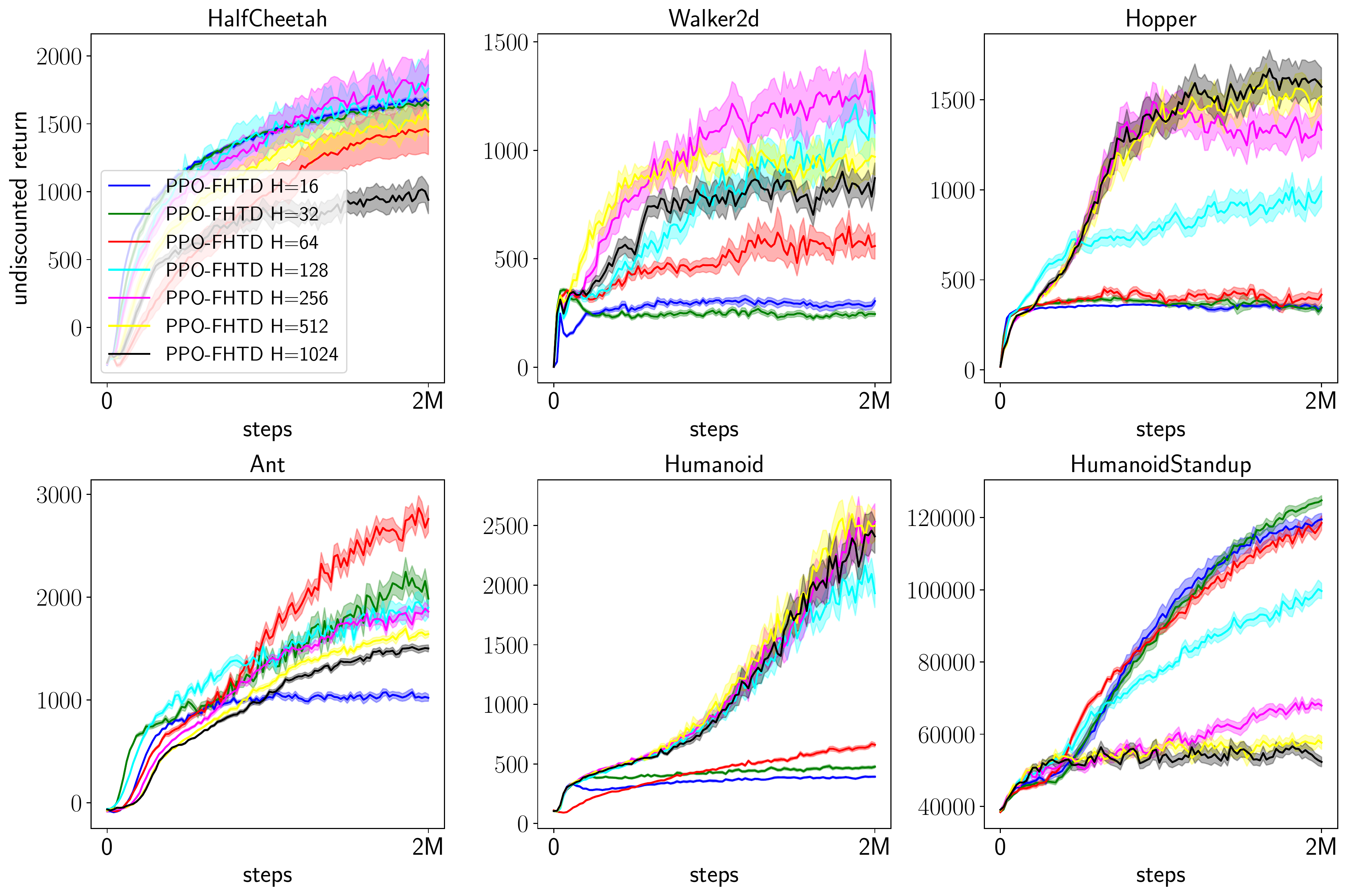}
\caption{PPO-FHTD with the first parameterization. The best $H$ and $\gamma_{\textsc{c}}$ are used for each game. The larger version of Figure~\ref{fig:ppo-fhtd-vs-ppo-td-ret}.}
\end{figure}
\begin{figure}[h]
\centering
\includegraphics[width=\linewidth]{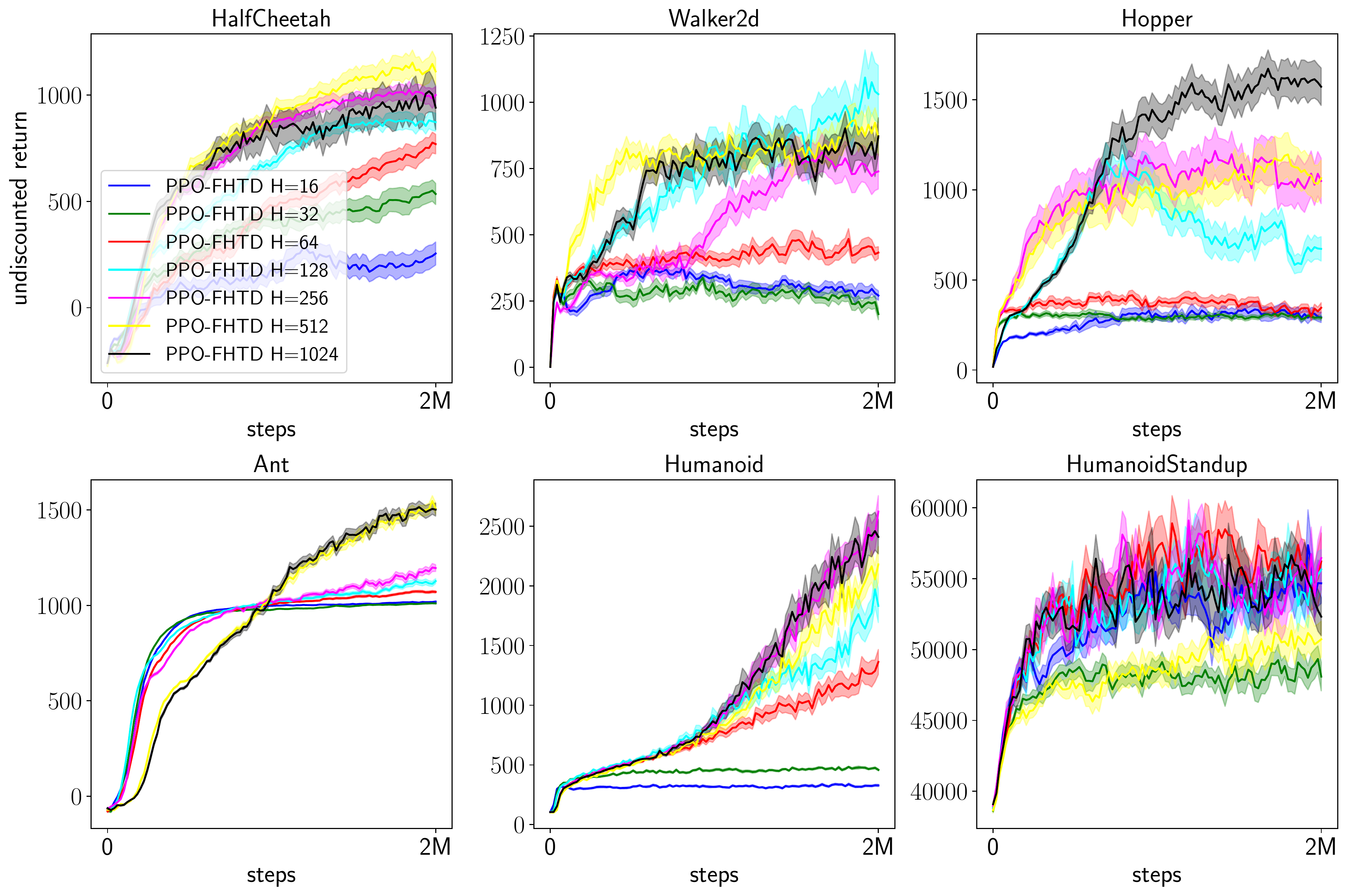}
\caption{PPO-FHTD with the second parameterization. The larger version of Figure~\ref{fig:ppo-active-fhtd-ret}.}
\end{figure}

\begin{figure}
\centering
\includegraphics[width=\linewidth]{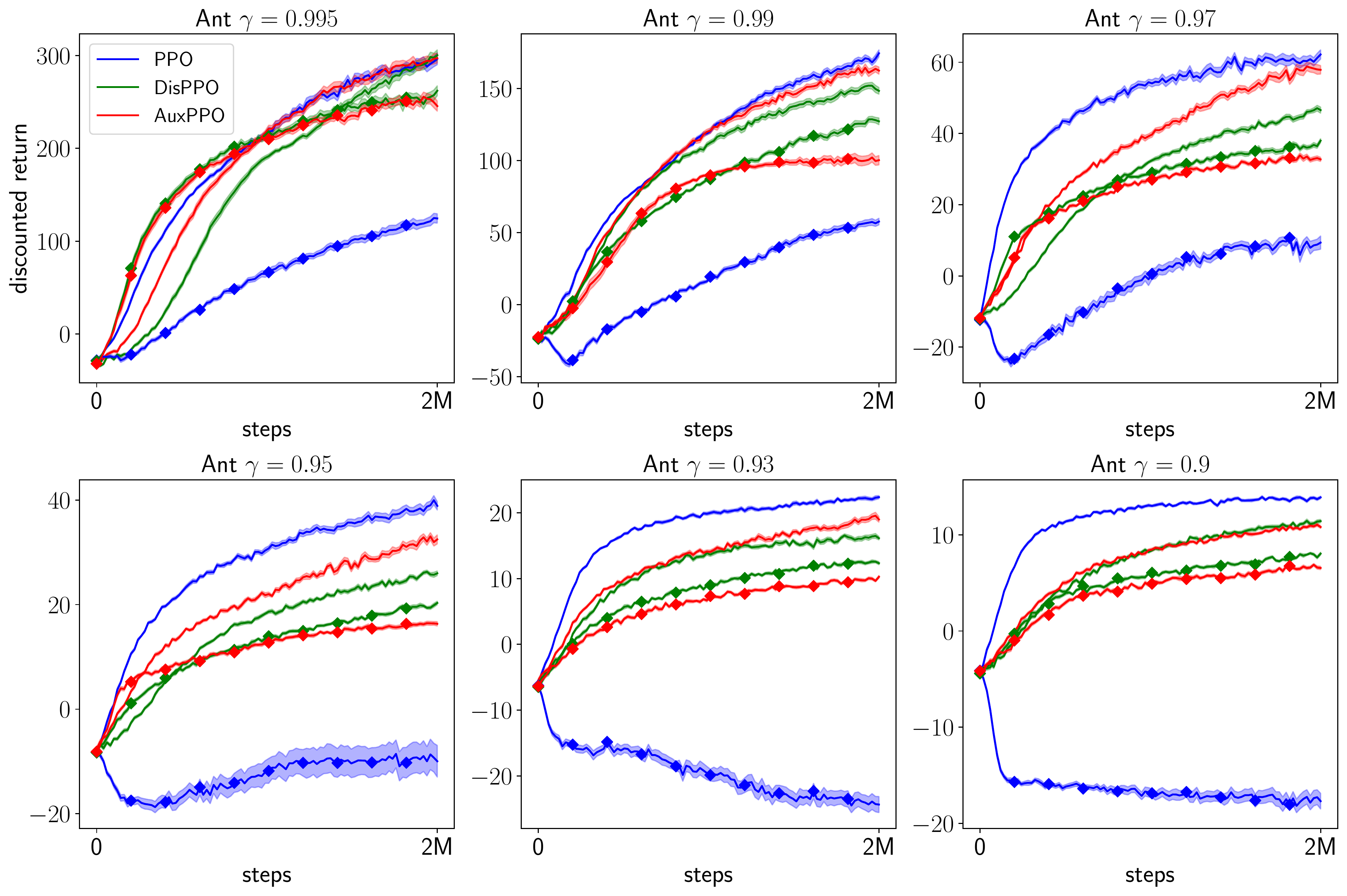}
\includegraphics[width=\linewidth]{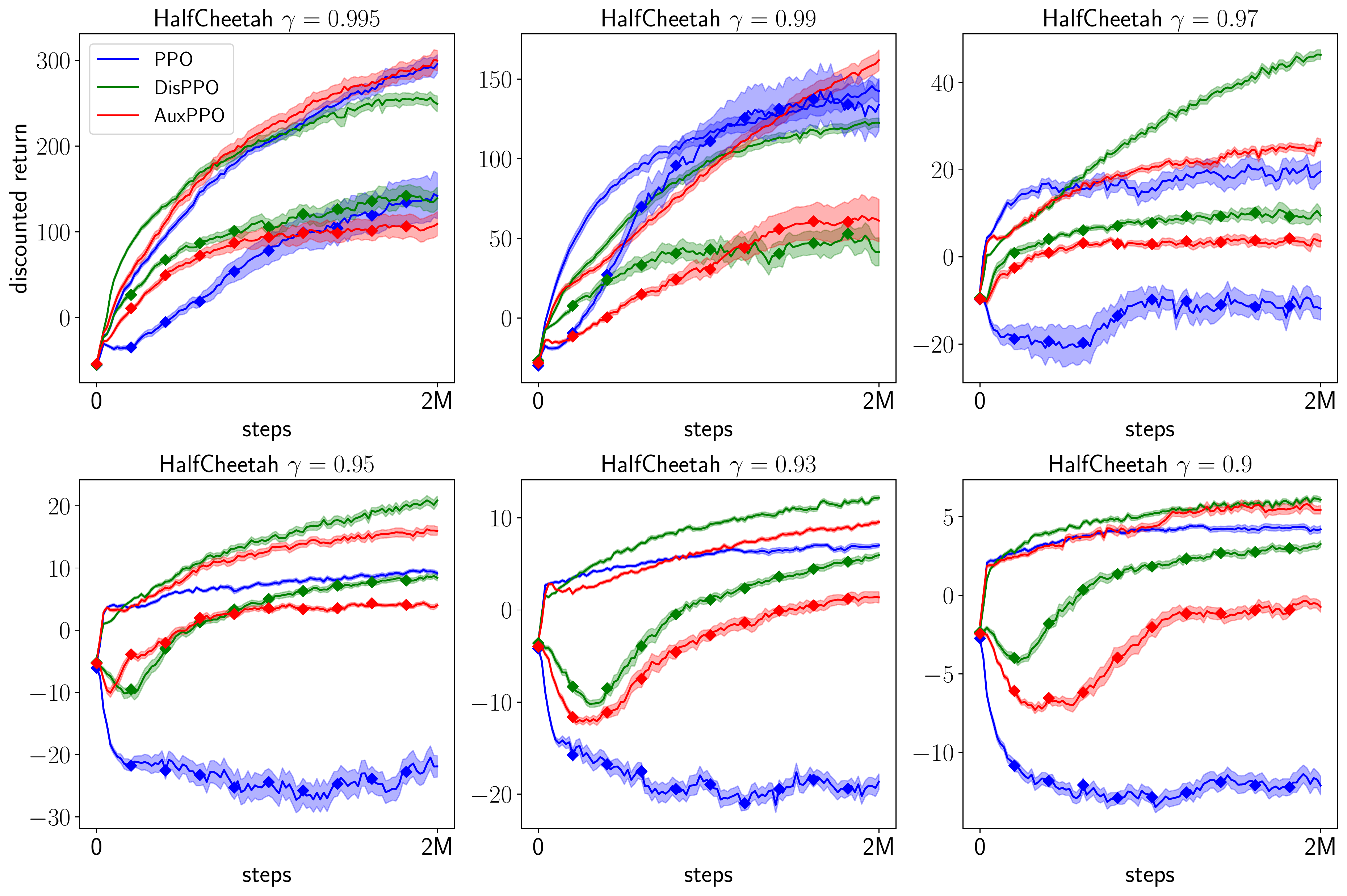}
\caption{Curves without any marker are obtained in the original \texttt{Ant} environment.
Diamond-marked curves are obtained in \texttt{Ant} with $r^\prime$.
The larger version of Figure~\ref{fig:aux_ppo}.
}
\end{figure}

\end{document}